\documentclass[a4paper,11pt]{article}

\usepackage[margin=1in]{geometry}
\usepackage[numbers]{natbib}
\usepackage[
  bookmarks,
  colorlinks=true,
  plainpages=false,
  hypertexnames=false,
  citecolor=red,
  linkcolor=blue,
  anchorcolor=red,
  urlcolor=blue
]{hyperref}
\usepackage{url}
\let\citet\cite

\usepackage{amssymb,amsmath,mathtools,amsfonts,amsthm,dsfont,bm}
\usepackage{float,xcolor,xspace,graphicx,appendix,enumitem,booktabs,multirow,longtable}
\usepackage{algorithm,algpseudocode}
\usepackage[normalem]{ulem}
\usepackage{tikz}
\usetikzlibrary{arrows.meta,shapes.geometric,positioning,calc,fit}
\usepackage{listings}

\makeatletter
\algrenewcommand\alglinenumber[1]{\footnotesize #1. }
\algrenewcommand\algorithmiccomment[1]{\hfill\textcolor{blue}{\(\triangleright\) #1}}
\makeatother

\usepackage{prettyref}
\usepackage{etoolbox}
\newlength{\sectiontocnumwidth}
\makeatletter
\patchcmd{\l@section}
  {\setlength\@tempdima{1.5em}}
  {\setlength\@tempdima{\sectiontocnumwidth}}
  {}{\PackageError{arxiv}{Unable to set TOC number width}{}}
\makeatother

\newrefformat{eq}{(\ref{#1})}
\newrefformat{sec}{Section~\ref{#1}}
\newrefformat{alg}{Algorithm~\ref{#1}}
\newrefformat{fig}{Figure~\ref{#1}}
\newrefformat{tab}{Table~\ref{#1}}
\newrefformat{rmk}{Remark~\ref{#1}}
\newrefformat{clm}{Claim~\ref{#1}}
\newrefformat{claim}{Claim~\ref{#1}}
\newrefformat{def}{Definition~\ref{#1}}
\newrefformat{assump}{Assumption~\ref{#1}}
\newrefformat{cor}{Corollary~\ref{#1}}
\newrefformat{lmm}{Lemma~\ref{#1}}
\newrefformat{prop}{Proposition~\ref{#1}}
\newrefformat{app}{\ref{#1}}
\newrefformat{thm}{Theorem~\ref{#1}}

\theoremstyle{plain}
\newtheorem{theorem}{Theorem}

\newtheorem{proposition}{Proposition}

\theoremstyle{definition}

\newtheorem{remark}{Remark}

\newtheorem*{theorem*}{Theorem}

\makeatletter
\newenvironment{breakablealgorithm}
  {%
   \begin{center}
     \normalfont\small
     \refstepcounter{algorithm}%
     \hrule height.8pt depth0pt \kern2pt%
     \renewcommand{\caption}[2][\relax]{%
       {\raggedright\textbf{\ALG@name~\thealgorithm} ##2\par}%
       \ifx\relax##1\relax
         \addcontentsline{loa}{algorithm}{\protect\numberline{\thealgorithm}##2}%
       \else
         \addcontentsline{loa}{algorithm}{\protect\numberline{\thealgorithm}##1}%
       \fi
       \kern2pt\hrule\kern2pt
     }
  }{%
    \kern2pt\hrule\relax
   \end{center}
  }
\makeatother

\newcommand{\algname}{ADP-HT}
\newcommand{\calF}{\mathcal{F}}
\newcommand{\E}{\mathbb{E}}
\newcommand{\1}{\mathbb{I}}
\newcommand{\R}{\mathbb{R}}
\newcommand{\clip}{\operatorname{clip}}
\providecommand{\argmax}{\operatorname*{arg\,max}}

\title{Pathwise-Budgeted LLM Routing: Online Allocation with $K$-Bit Tail-Cost Learning}
\date{}
\author{
  Hanzhang Qin,
  Sophie H.\ Yu,
  Yuewei Zhang, and Zhi-Hai Zhang\thanks{
  Authors are listed in alphabetical order.\\
  \hspace*{\parindent}H. Qin is with the Department of Industrial Systems Engineering and Management, National University of Singapore, \texttt{hzqin@nus.edu.sg}.
  S.\ H.\ Yu is with The Wharton School of Business, University of Pennsylvania, Philadelphia PA, USA,  \texttt{hysophie@wharton.upenn.edu}.
  Y. Zhang is with the Department of Industrial Engineering, Tsinghua University, Beijing, China, \texttt{zyw23@mails.tsinghua.edu.cn}.
  Z.H Zhang is with the Department of Industrial Engineering, Tsinghua University, Beijing, China, \texttt{zhzhang@tsinghua.edu.cn}. \\\hspace*{\parindent}Y. Zhang is the corresponding author.}
}
\begin{document}
\maketitle

\begin{abstract}
Firms using large language models under fixed budgets must decide which model to call and whether to retry a failed task. Because a call's charge is revealed only after completion, one long response can exhaust the remaining budget and leave later tasks unserved. Our objective is to maximize the number of tasks solved before service stops. We develop a contextual bandit policy, \algname{},  which learns each model's task-specific success probability and constructs conservative cost estimates from coarsened response-length labels on a quantile grid with $2^K$ intervals. The policy weighs predicted success against these costs, adjusting the cost penalty to realized spending and the budget available per remaining task. Under calibration, local regularity, and grid-coverage conditions, we bound the excess allowance in the policy's conservative cost estimates. At a fixed sample size, increasing $K$ reduces the discretization term but increases the estimation allowance. At the tightest tested budgets, ADP-HT solves more than twice as many MATH500 and GPQA tasks as the best implemented baseline. It achieves the highest mean cumulative score at ten of fourteen RouterBench budgets. On LiveCodeBench, it outperforms a deployed commercial router in both default and cost-saving modes at the three lowest test budgets.
\end{abstract}


\setcounter{tocdepth}{2}
\tableofcontents

\section{Introduction}
\label{sec:intro}

Large language models (LLMs) are now used at scale in software development, customer service, and other knowledge-intensive work. As these applications expand, finite budgets for model use have become a central operating constraint. Developer tools and model platforms typically define these budgets over rolling or monthly usage periods and meter them in dollars or tokens \citep{github2026copilotdocs,anthropic2026pricing}. Providers differ sharply in capability and price, so the same budget can support materially different amounts of service depending on which models are used. A call's final charge is also uncertain at admission, since it depends in part on the length of the response. Together, finite budgets, price dispersion, and uncertain call costs create an operational opportunity for routing. By matching each task to a capable model whose cost fits the remaining budget, users complete more work before the budget is exhausted.

Much of the literature on LLM routing treats model choice as a quality--cost trade-off for the current request. Cascades decide whether to escalate an inexpensive response, while supervised routers learn which model is likely to deliver sufficient quality at an acceptable expected cost \citep{chen2023frugalgpt,madaan2024automix,ding2024hybridllm,ong2025routellm,somerstep2025carrot}. This expected-value perspective is natural when requests can be considered independently or spending is controlled on average. However, a finite pathwise budget in most LLM services is enforced based on cumulative realized spending along the actual sequence of calls, rather than on expected spending. This stopping rule changes the consequences of call-cost uncertainty. A call's realized charge becomes known only after completion, and unusually long responses are difficult to anticipate. One call can therefore exhaust the remaining budget and end service for every later task. Its consequence includes both the current charge and the loss of future service. Optimizing expected quality against expected cost request by request does not control this stopping risk.

Consider a developer with a \$100 monthly budget for model calls on OpenRouter. Coding tasks arrive one at a time, from a short unit-test fix to a difficult multi-file change. For each task, the developer chooses among models including GPT-5.6 Sol, Claude Opus 5, Kimi K3, GLM 5.2, and others. The models differ in price and in the kinds of tasks they solve well \citep{openrouter2026models,openrouter2026pricing}. Only after a call finishes does the developer observe its total token count and exact provider charge. If the response fails, the developer must decide whether another call, now carrying the failed exchange in a longer prompt, is worth its cost or whether to abandon the task and preserve budget for later work. Each selected call reveals its model's outcome and response length, whereas the outcomes of unselected models remain unknown. The challenge is to learn which model fits each task while preserving enough budget to complete as many later tasks as possible.

The same operating structure arises when firms assign API budgets to employees, when an LLM customer-service system handles cases of different difficulty, and when a metered AI service allocates model access across client requests. We focus on settings with three features. Tasks are processed online under a pathwise budget. Call cost is unknown at admission and can vary substantially in its upper tail. Success and cost profiles must be learned from selected calls. These features lead to our central operations question: \textit{how should an operator route and retry a sequential task stream to maximize the number of tasks completed before a finite LLM budget stops service?}

To answer this question, we formulate LLM routing as a \emph{$K$-bit Pathwise Contextual Bandits with Knapsacks} problem and develop \algname{} (Adaptive Dual-Pace Hybrid-Threshold), an online policy that weighs each call's chance of solving the current task against the future service its uncertain cost puts at risk. Our contributions are fourfold.

First, we formalize LLM routing under a budget with a halt-after-crossing rule, treating each model as an arm. A successful response removes a task from the stream, a failed task may be retried or abandoned, and the first call whose charge crosses the budget ends service. The objective is to maximize the number of tasks solved before this stopping time. LLM's response lengths are heavy-tailed and difficult to predict \citep{wang2026prod}, and a single long response may exhaust the remaining budget and terminate service for all future tasks. For such a halt-after-crossing budget, the cost needs to be forecast in a risk-aware manner rather than an expected point estimate sense. We therefore introduce a $K$-bit cost-shape channel, which discretizes response length into $2^K$ intervals and estimates the exceedance probability at each interval boundary. This changes the cost prediction problem from estimating the exact response length to estimating whether the response length would be longer or shorter, which is more robust and reliable. A larger $K$ improves the resolution of prediction but may increase the estimation difficulty, which introduces a trade-off between resolution and precision. This structure extends the random-cost, halt-after-crossing formulation of \citet{cayci2020budget} from repeated arm pulls to an online task stream with per-task value, router-controlled retries, and $K$-bit cost prediction.

Second, we give an implementable online policy for choosing models and retries. A hybrid reward learner combines a shared low-dimensional component with model-specific adjustments to form an optimistic solve probability. Threshold-survival heads use the quantized targets to estimate a pre-call cost bound at a chosen tail tolerance. The selector subtracts that shadow-priced cost bound from the optimistic solve probability, so an expensive model remains eligible when its chance of solving the task compensates for the budget it would consume. A dual-pace controller adjusts the shadow price from realized spending and smooths the pacing response through recent average costs. After a failure, the retry gate raises its margin with the attempt index and spending beyond the revised task allowance.

Third, we characterize the $K$-bit trade-off between resolution and precision and show that its discretization and estimation terms are jointly order-tight under the stated confidence calibration. The implemented grid divides a core range containing the more common response lengths into quantile-based intervals and reserves one tail interval for the remaining long responses. Our main theorem separates the excess cost allowance arising from grid coarseness, warm-data placement of the grid edges, and simultaneous estimation of the threshold-crossing probabilities. More bits reduce the discretization term but increase the estimation allowance at a fixed sample size. The result applies when the target response-length threshold, including its estimation allowance, remains in the core. The same argument extends to other fixed grids through their local edge spacing. We also ask how the value of tail protection changes with the remaining workload even when cost distributions are known, by solving a two-model specialization with one call per task. At a fixed budget, the optimal first choice can move from the risky model to the predictable model and then back as the number of remaining tasks increases. Predictable calls protect future service, but their higher ordinary cost can displace useful calls when many tasks remain.

Fourth, three numerical experiments evaluate the policy, from benchmark task streams to a deployed commercial router. At the tightest tested budgets, \algname{} solves an average of $216.2$ \textsc{MATH500} problems and $85.0$ \textsc{GPQA} questions, more than twice the best implemented online comparator in each dataset. On \textsc{RouterBench}, the implemented \algname{} policy leads in the descriptive ranking at all eight tight budgets and ranks first in ten of fourteen budgets overall. In the commercial experiment, \algname{}'s test phase solves 682 problems while incurring $\$0.8084$, whereas default AutoRouter's complete run solves 623 for $\$22.86$.

The rest of the paper is organized as follows.
Section~\ref{sec:related} reviews the related literature to situate our formulation among bandits with resource constraints, tail-aware learning, and cost-aware LLM routing.
Section~\ref{sec:model} defines the formal model including the task stream, exact budget ledger, and stopping rule.
Section~\ref{sec:policy} develops \algname{}, which prices every call and retry against the budget it puts at risk and characterizes the $K$-bit trade-off in pre-call cost precision.
Section~\ref{sec:experiments} evaluates the policy in three experiments, from live routing with retries to cross-task allocation on a public benchmark and a comparison with a deployed commercial router.
Section~\ref{sec:conclusion} concludes.

\paragraph{Notation.} Throughout the paper, we write $[n]:=\{1,\ldots,n\}$ and $\1\{\cdot\}$ for the indicator of an event. For $l\le u$, $\clip_{[l,u]}\{v\}:=\min\{u,\max\{l,v\}\}$ truncates $v$ to $[l,u]$. A tilde marks a random potential outcome and the same symbol without a tilde its realized value, while a hat marks a quantity estimated from data. Finally, $\widetilde O(\cdot)$ hides polylogarithmic factors.

\section{Literature Review}
\label{sec:related}

Our work draws from and contributes to three streams of literature, which we connect by examining how a depleting pathwise budget and coarsened tail-cost information jointly shape model-choice and retry decisions.

\subsection{Online Resource Allocation under Random Consumption}

Online resource allocation provides the primary operations perspective for our problem. In revenue management, bid prices express the opportunity cost of consuming capacity now rather than reserving it for later demand \citep{talluri1998analysis,topaloglu2009capacity}, and online-allocation policies extend this logic by updating resource prices as demand and consumption unfold \citep{balseiro2023bestofmany}. The same logic applies to an LLM budget. Every admitted call spends capacity that could otherwise serve a later task, and our policy prices each call against that displaced service.

Bandits with knapsacks bring learning into this opportunity-cost framework. Canonical models combine selected-arm feedback with a finite resource \citep{tranthanh2012knapsack,badanidiyuru2018bwk}. Contextual and regression-oracle variants allow reward and consumption to depend on observed covariates \citep{agrawal2016linearcbwk,slivkins2024cbwlc,zhalechian2022personalized}. Related learning problems arise in inventory control and reusable-resource pricing \citep{tang2026upgrading,jia2022linear}. Random-cost formulations allow jointly distributed costs and rewards that are observed after selection \citep{cayci2020budget}. Our formulation retains the finite budget, random consumption, and stopping upon depletion, and applies them to an online stream of tasks.

Within this stream, \citet{cayci2020budget} provides a close starting point. Each pull generates a random cost--reward pair, play ends when cumulative cost first exceeds the budget, and the last pull's reward is retained. Their model allows signed and potentially unbounded costs and rewards. We focus on nonnegative token charges, stop when spending reaches or exceeds the budget, and retain the final response's reward. We add three operational features to this post-call accounting. First, the allocation unit is an online stream of distinct tasks rather than repeated arm pulls, so value is earned only when a task is completed and a verified task leaves the stream. Second, continuation is itself a decision. After a failed call, the policy chooses whether another call should compete with unserved tasks for the shared budget. Third, the cost predictor uses quantized response-length targets to estimate the upper tail of the next call's cost. This tail matters for both model choice and continuation because a large realized charge can end service for later tasks.

\subsection{Tail-Sensitive Learning with Coarsened Cost Feedback}

Risk-sensitive and safety-constrained bandits protect different aspects of sequential performance. Conservative policies require cumulative performance to remain above a prescribed fraction of a baseline, with guarantees in expectation or with high probability \citep{wu2016conservative}. Related formulations address anytime cumulative constraints \citep{liu2021pessimistic} and chance constraints on resource use \citep{cohen2019overcommitment,lu2021nonprofit}. Gambler and survival bandits model termination at a ruin threshold for accumulated reward or wealth \citep{perotto2021gambler,riou2024survival}. Quantile forecasts describe uncertainty in service outcomes \citep{arora2023probabilistic}, while quantile-bandit and sequential-estimation methods estimate reward quantiles and identify arms by their quantile performance \citep{zhang2021quantilebandits,howard2022quantile}. Our focus is the conditional upper tail of the next call's cost.

A second stream shows that aggressively coarsened feedback can support effective shape learning, which is operationally relevant when the exact cost is unknown at admission and hard to predict. Communication-constrained bandits quantize rewards into a few bits while preserving learning performance \citep{hanna2022fewbits}, and one-bit quantile bandits use adaptive threshold queries to identify the arm with the highest reward quantile \citep{lau2025onebitquantile}. We bring this coarsened-feedback idea to cost-tail estimation. The exact provider charge updates the pathwise ledger, while the cost learner receives only which of $2^K$ fixed response-length intervals contains the call. The resolution parameter $K$ thereby becomes an information-design parameter, since a finer mesh localizes the conditional tail more closely but enlarges the set of thresholds estimated simultaneously, and our analysis characterizes this finite-sample trade-off  for the algorithm we propose.

\subsection{Cost-Aware and Sequential LLM Routing}
LLM routing methods differ in whether they enforce a budget through realized cumulative spending across tasks and whether they jointly model uncertain call costs and retries. Cascades such as FrugalGPT and AutoMix escalate from an inexpensive model when its response is judged insufficient \citep{chen2023frugalgpt,madaan2024automix}. HybridLLM, RouteLLM, and CARROT instead learn a request-level choice along a quality--cost frontier \citep{ding2024hybridllm,ong2025routellm,somerstep2025carrot}, and RouterBench supplies an offline environment for comparing such routers \citep{hu2024routerbench}. Adaptive methods reach further. PILOT allocates estimated costs over a known query set \citep{panda2025pilot}, \citet{poon2026multillm} control up to $H$ calls within a query and screen models by confidence bounds on mean cost, ParetoBandit paces an average per-request cost ceiling over an open-ended stream \citep{tabernermiller2026paretobandit}, ACQB uses user retrials as implicit feedback for joint routing and scheduling \citep{bae2026acqb}, and SeqRoute and MTRouter learn offline from logged trajectories how a model choice bears on the terminal outcome of a session \citep{xu2026seqroute,zhang2026mtrouter}. \citet{bu2026oscar} develop OSCAR and analyze which LLM configurations to call and when to stop while searching for the next simulator-certified improvement. Related stopping decisions arise in contextual search \citep{belloni2026pandora} and LLM-based sequential testing \citep{li2026sequential}. Our decision unit is instead a shared task-stream budget that binds pathwise by halting, so a call's consequence includes the future service its realized cost displaces, and the exact cross-task ledger, the tail-cost bound for the next call, and router-controlled retries must be modeled jointly. The router also learns online from chosen-arm feedback on verifier-checked tasks rather than offline from logs.

\section{Problem Statement}
\label{sec:model}

We consider a budget holder who calls LLMs directly under a single monetary or token amount fixed for the period, such as the developer of Section~\ref{sec:intro}. Tasks arrive online, one at a time, and each is served with one of the available models. After a failed response, the decision maker may retry with the failed exchange added to the prompt or leave the task for manual work. The objective is to maximize the number of tasks completed before the budget is exhausted. Each call both draws down the budget and generates data about the model that was called, so the exploration--exploitation trade-off is central to the problem. A model's output is observed only after the call returns, and its length, the part of the charge that is unknown at admission, is difficult to predict accurately in advance, even for a fixed prompt on a fixed model \citep{wang2026prod}. The remainder of this section fixes the primitives of the problem, namely the task stream, the information available before a call, the pathwise ledger, and the rule that stops service when spending reaches or crosses the budget.

\subsection{Tasks, Models, and Information}

We model this problem in a bandit framework. Let $\mathcal M$ be the finite set of available LLMs and regard each one as an arm, so that $M=|\mathcal M|$ is the number of arms. The total number of tasks $T$ and the total budget $B$ are known in advance, but the contents and difficulty of future tasks are unknown until they arrive. The tasks $q_1,\ldots,q_T$ arrive online, one at a time, and are served under this budget. For each task, the decision maker selects and calls an LLM in an attempt to solve it. We call the policy that makes these model and retry choices the \emph{router}. A task can receive at most $H\ge1$ calls and leaves the stream after its first successful response, after the router abandons it, or after all $H$ attempts have been used. For bookkeeping, we order the $S=TH$ potential task--attempt steps lexicographically and index them by $s=(t-1)H+h\in[S]$, where $t\in[T]$ is a task and $h\in[H]$ an attempt. Conversely, $(t_s,h_s)$ denotes the task and attempt associated with step $s$, with $t_s=\lceil s/H\rceil$ and $h_s=s-(t_s-1)H$. Let $G_s\in\{0,1\}$ indicate whether step $s$ executes a call. For a first attempt when there is still remaining budget, we always have $G_s=1$. While for a retry, $G_s=1$ only if the router chooses to call rather than abandon and move to next task. For an abandoned or solved task, $G_u=0$ for every later step $u>s$ with $t_u=t_s$. And we write $a_s\in\mathcal M$ for the arm called when $G_s=1$. When $H=1$, the problem contains one model choice per task and no retry decision.

\paragraph{State and information.}
Before step $s$, the router observes a state $x_s$ in a state space $\mathcal X$, a context summary constructed from known information, including the current prompt, the failed responses already obtained for task $t_s$, the attempt index $h_s$, and the budget ledger. Let $\calF_s$ be the $\sigma$-field generated by all states, decisions, outcomes, and ledger values realized through step $s-1$ together with the current state $x_s$, so that $\calF_s$ is the information available before the response, and the sequence $\{\calF_s\}_{s\in[S]}$ is the \emph{pre-outcome filtration}. Every pre-call quantity, including the state, the budget, the decision to call, and each selection score, is $\calF_s$-measurable, whereas the response and its cost are not. We take $x_s$ to be sufficient for the response, in the sense that the solve probability and the response-length distribution of a called arm depend on the past only through $x_s$, so the conditional means and survival probabilities used in Section~\ref{sec:policy} are functions of $(a,x_s)$. The router's estimators, shadow price, and pacing multiplier, by contrast, depend on the whole realized path, so decisions are required to be $\calF_s$-measurable rather than functions of $x_s$ alone. The policy later uses two deterministic views of the observed state, a reward feature $x_s^r\in\R^{d_r}$ and, for each candidate arm, a cost feature $x_s^c(a)\in\R^{d_c}$ where we use $r$ and $c$ as upper or lower indices to denote features for reward and cost, respectively.

\paragraph{Outcomes and feedback.}
If a candidate arm $a$ is called in state $x_s$, it produces a solve indicator $\tilde r_s(a,x_s)\in\{0,1\}$ and an output length $\tilde\ell_s^{\rm out}(a,x_s)\in[0,\ell_{\max}^{\rm out}]$, where $\ell_{\max}^{\rm out}$ is a known upper bound on response length (GPT-5.6 Sol, for instance, caps completions at $128{,}000$ tokens \citep{openai2026models}). We assume the verifiable binary success feedback becomes available after the call. For example, an execution grader can run code tests for a coding task, or a verifier can compare a mathematical answer with its reference answer. Providers charge by the token, so the call costs
\begin{align}
\tilde c_s(a,x_s)=\rho^{\rm in}_{a}\ell_s^{\rm in}(a)+\rho^{\rm out}_{a}\tilde\ell_s^{\rm out}(a,x_s),
\label{eq:priced-cost}
\end{align}
where $\ell_s^{\rm in}(a)$ is the prompt length in tokens under arm $a$'s own tokenizer, known before the call, and $\rho_a^{\rm in},\rho_a^{\rm out}>0$ are the known input and output prices per token. Only the selected arm reveals outcomes. Its realized values are $r_s$, $\ell_s^{\rm out}$, and $c_s=\rho^{\rm in}_{a_s}\ell_s^{\rm in}(a_s)+\rho^{\rm out}_{a_s}\ell_s^{\rm out}$. The input side of the charge is therefore known at admission, whereas the output side is not.

\subsection{Decision Process}
\label{sec:process}
\begin{figure}[!htbp]
\centering
\resizebox{0.7\linewidth}{!}{%
\begin{tikzpicture}[
  yscale=0.80,
  >={Stealth[length=2.2mm,width=1.6mm]},
  font=\fontsize{9}{10.8}\selectfont,
  procbox/.style={draw,rounded corners=2pt,align=center,inner sep=4pt,text width=5.6cm},
  sidebox/.style={draw,rounded corners=2pt,align=center,inner sep=4pt,text width=4.2cm},
  stopbox/.style={draw,thick,rounded corners=2pt,fill=black!6,align=center,inner sep=4pt,text width=4.2cm},
  testbox/.style={draw,diamond,aspect=2.4,align=center,inner sep=1pt},
  flowline/.style={->,semithick},
  edgelab/.style={font=\fontsize{8}{9}\selectfont,inner sep=1.5pt,fill=white},
  stage/.style={draw,dashed,black!60,rounded corners=4pt,inner sep=6pt},
  stagelab/.style={font=\fontsize{8}{9}\selectfont\bfseries,black!70,inner sep=2pt}
]
\node[procbox,text width=4.2cm] (start) at (0,0.9) {Initialize $C\leftarrow0$ and $t\leftarrow1$};
\node[procbox,text width=4.6cm] (arrive) at (0,-0.6) {Task $q_t$ enters service\\ with attempt $h=1$};
\node[procbox,text width=6.4cm] (obs) at (0,-2.3)
  {\textbf{(i) Observe.} State $x$: prompt of $q_t$,\\ failed responses, attempt $h$;\\ remaining budget $R=B-C$};
\node[testbox] (gate) at (0,-3.95) {$R>0$?};
\node[testbox] (firstq) at (0,-5.35) {$h=1$?};
\node[procbox] (call) at (0,-6.8)
  {Select an arm $a\in\mathcal M$ and call it};
\node[procbox] (real) at (0,-8.3)
  {\textbf{(iii) Realize.} Observe $(r,\ell^{\rm out},c)$;\\ ledger $C\leftarrow C+c$};
\node[testbox] (settle) at (0,-10.1) {$r=1$ or $h=H$?};
\node[procbox,text width=4.6cm] (inc) at (0,-11.7)
  {$h\leftarrow h+1$; append the failed\\ response to the state};
\node[stopbox,text width=4.6cm] (stopped) at (6.4,-2.3)
  {\textbf{Stop.} No further call is admitted;\\ $q_t$ and all later tasks are lost};
\node[testbox] (retryq) at (6.4,-5.35) {Retry?};
\node[sidebox] (leave) at (6.4,-10.1)
  {Task $q_t$ leaves the stream, solved and credited or unsolved; then $t\leftarrow t+1$, and the stream ends after $q_T$};
\node[stage,fit=(gate)(firstq)(retryq)(call)] (grpii) {};
\node[stagelab,anchor=north west] at ([xshift=2pt,yshift=-2pt]grpii.north west) {(ii) Decide};
\node[stage,fit=(settle)(inc)(leave)] (grpiv) {};
\node[stagelab,anchor=north west] at ([xshift=2pt,yshift=-2pt]grpiv.north west) {(iv) Transition};
\draw[flowline] (start) -- (arrive);
\draw[flowline] (arrive) -- (obs);
\draw[flowline] (obs) -- (gate);
\draw[flowline] (gate) -- node[edgelab,right]{yes} (firstq);
\draw[flowline] (gate.east) -- node[edgelab,above]{no} (6.4,-3.95) -- (stopped.south);
\draw[flowline] (firstq) -- node[edgelab,right]{yes} (call);
\draw[flowline] (firstq) -- node[edgelab,above]{no} (retryq);
\draw[flowline] (retryq.south) |- node[edgelab,pos=0.15,right]{yes} (call.east);
\draw[flowline] (retryq.east) -- node[edgelab,above]{no: abandon} (9.2,-5.35) -- (9.2,-10.1) -- (leave.east);
\draw[flowline] (call) -- (real);
\draw[flowline] (real) -- (settle);
\draw[flowline] (settle.east) -- node[edgelab,above]{yes} (leave.west);
\draw[flowline] (settle) -- node[edgelab,right]{no} (inc);
\draw[flowline] (inc.west) -- (-3.6,-11.7) -- (-3.6,-2.3) -- (obs.west);
\node[edgelab,rotate=90] at (-3.6,-6.9) {next attempt of task $q_t$};
\draw[flowline] (leave.south) -- (6.4,-12.9) -- (-5.1,-12.9) -- (-5.1,-0.6) -- (arrive.west);
\node[edgelab,rotate=90] at (-5.1,-6.6) {next task, with the ledger $C$ carried forward};
\end{tikzpicture}}
\caption{Decision flow of the task stream under the budget $B$.}
\label{fig:decision-flow}
\end{figure}
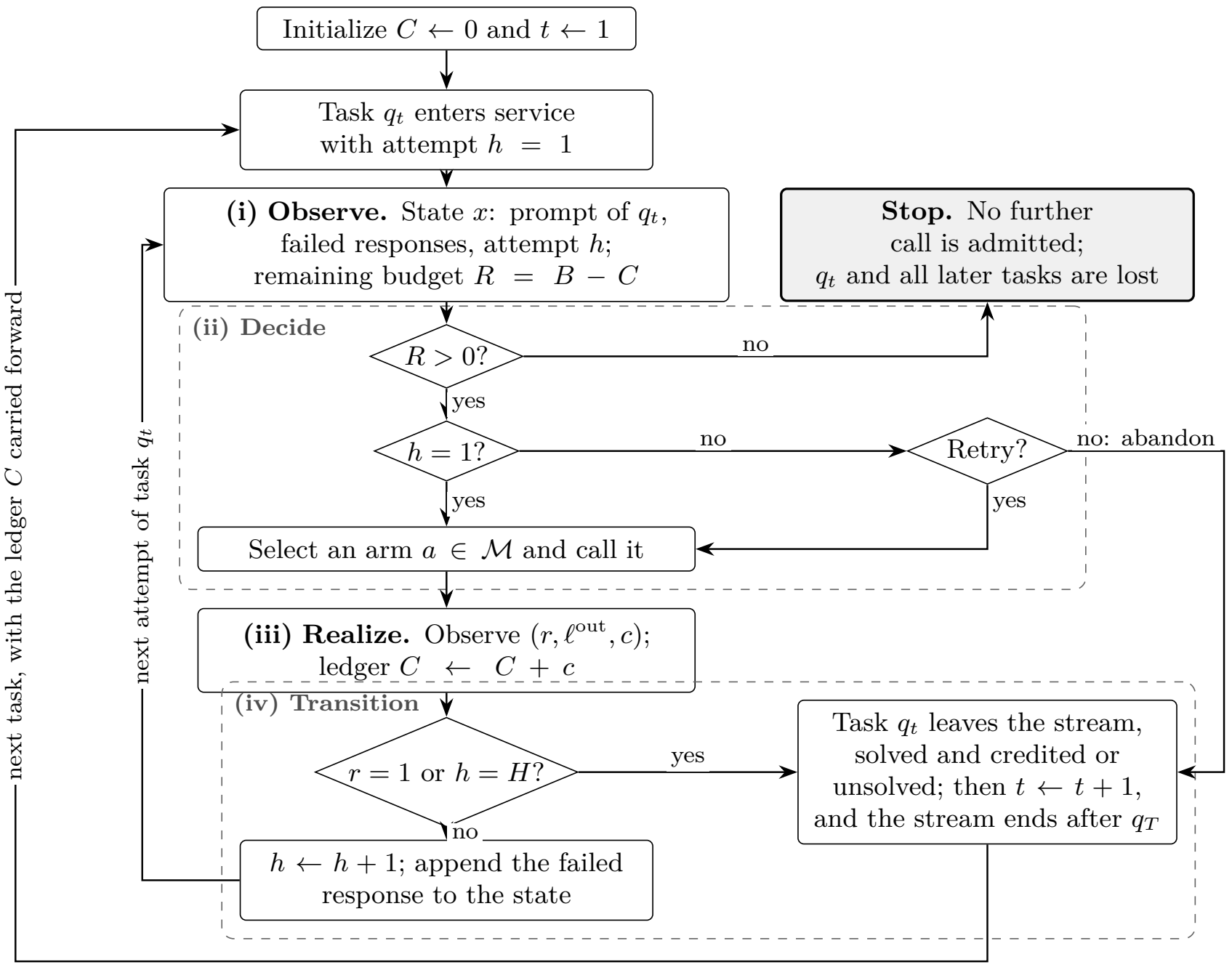

Let $C_{s-1}=\sum_{u<s}G_uc_u$ be the exact spend before step $s$ and $R_s=B-C_{s-1}$ the remaining budget. Each step proceeds in four stages.
\begin{enumerate}
\item[(i)] \textbf{Observe.} The router reads the state $x_s$, including the prompt, retry history, and current ledger.
\item[(ii)] \textbf{Decide.} The step is \emph{live} if task $t_s$ is unsolved and not abandoned and $R_s>0$. At a live first attempt the router selects an arm $a_s\in\mathcal M$ and calls it. At a live retry it either calls an arm or abandons the task. Set $G_s=1$ for a call and $G_s=0$ otherwise. A task that has been solved or abandoned is no longer live, so $G_u=0$ at every later step $u>s$ with $t_u=t_s$.
\item[(iii)] \textbf{Realize.} If $G_s=1$, the selected arm returns $(r_s,\ell_s^{\rm out},c_s)$ and the ledger becomes $C_s=C_{s-1}+c_s$. If $G_s=0$, no call is made, $c_s=0$, and $C_s=C_{s-1}$.
\item[(iv)] \textbf{Transition.} If $G_s=1$ and $r_s=1$, task $t_s$ is solved and the stream advances. If $G_s=1$, $r_s=0$, and $h_s<H$, the failed response is appended to the state used for the next attempt. A failure at $h_s=H$ leaves the task unsolved and advances the stream, as does an abandonment decision taken at stage~(ii).
\end{enumerate}
Thus every executed step includes model selection, while admission control begins with the first retry. Figure~\ref{fig:decision-flow} draws the resulting process at the task--attempt level, with step subscripts suppressed. The router decides from pre-call information, observes the selected model only after execution, and then updates both the state and the budget, so that one realized charge tightens every later admission.

\subsection{Pathwise Budget and Stopping Rule}
\label{sec:stopping}

Executed charges accumulate in the ledger $C_s=\sum_{u\le s}G_uc_u$. A call can be admitted whenever $R_s>0$, but its realized charge may exceed the remainder. Service stops after the first call that reaches or crosses the budget, at
\begin{align}
\tau_B=\inf\{s\ge1:C_s\ge B\},
\label{eq:tau}
\end{align}
with $\tau_B=\infty$ when no crossing occurs. The crossing response is observed and receives credit if it solves the current task, but $G_u=0$ for every later step $u>\tau_B$. This \emph{halt-after-crossing} rule therefore limits future admissions rather than truncating a response already in service.

This pathwise rule makes the consequence of an expensive call depend on its position in the task stream. Beyond the current charge, its loss includes the later tasks the charge removes from service, so the budget is a one-dimensional, absorbing capacity constraint rather than an average-cost penalty applied independently to each request.

\subsection{Objective}

Let $\Pi_{\rm na}$ be the class of nonanticipating policies, those whose step-$s$ decisions are $\calF_s$-measurable. The router maximizes the expected number of distinct tasks solved before service stops, with $\E_\pi$ denoting expectation under the policy $\pi$:
\begin{align}
\max_{\pi\in\Pi_{\rm na}}\ \E_\pi[\textsc{Solved}(\pi)],\qquad
\textsc{Solved}(\pi)
=\sum_{t=1}^{T}\1\{\exists\,s\le\min\{S,\tau_B\}:\ t_s=t,\ G_s=1,\ r_s=1\}.
\label{eq:objective}
\end{align}
This objective values each unit of budget through the tasks it keeps serviceable, rather than through a per-call quality--cost ratio.

\section{Methodology}
\label{sec:policy}
We develop \algname{}, an online policy for LLM routing, which is motivated by the bid-price principle of online resource allocation. Consider a budgeted bandit problem. In each round, the decision maker selects an arm, receives a reward $r$, and consumes $c$ units of a limited resource. Resource spent on the current pull is no longer available in later rounds. Then each arm's value can be measured by the reward it produces minus the opportunity cost of the resource it consumes, which can be expressed as
\begin{align}
Q=r-\lambda c, \label{eq:score}
\end{align}
where $\lambda$ is the \emph{shadow price}, the opportunity cost of consuming one unit of budget. And $Q$ can be viewed as the net value of the action, so the decision maker should select the arm with the highest $Q$.

However, this score is not directly available in LLM routing. In this setting, each arm corresponds to a candidate LLM. The reward $r$ records whether its call solves the current task, with $r=1$ for success and $r=0$ otherwise, and $c$ is the budget charged for the call, but the router does not know either exactly before a call is made. For example, even when the prompt length and token prices are known, the number of output tokens is unknown until the response is generated, and the exact charge cannot be determined in advance. So it has to learn the solve probability from selected calls and estimate the cost before the response returns. Because one long response can exhaust the remaining budget, the cost term should take this risk into account. The shadow price should also track realized spending adaptively to better utilize the remaining budget. And for tasks that fail on the first attempt, the router needs to decide whether to retry the same task or preserve the budget for later tasks. The rest of this section builds \algname{} based on the score \eqref{eq:score} with modifications for these challenges.

\algname{} supplies these quantities in four modules designed for LLM routing. For the reward (see Section \ref{sec:reward-module}), a hybrid learner combines a shared task-difficulty component with a model-specific adjustment and reports an upper confidence bound (UCB) on the solve probability. For the cost (see Section \ref{sec:cost-module}), we utilize a $K$-bit prediction of the response length to construct a one-sided pre-call cost bound that covers the tail of the response-length distribution. It is difficult to predict the exact output tokens for a LLM, but it is much easier to predict whether the response will be short, medium, or long. This motivates the $K$-bit threshold-survival heads, which can help the router predict more reliably with a loss of resolution, introducing a trade-off between resolution and precision. The shadow price is controlled by a dual-pace mechanism (see Section \ref{sec:capacity-module}). Its fast component updates the price directly from the latest call's charge, while its slow component updates a pacing multiplier from an exponential moving average of recent charges. These components adjust the cost penalty and the current task's spending reference in response to realized spending and remaining capacity. Finally, the retry gate (see Section \ref{sec:retry-module}) decides whether a failed task should continue competing for capacity, using the current pre-call score of the selected candidate and a margin computed after the previous failure. The margin rises with retry depth and with spend beyond the revised task allowance. The following subsections describe these modules, and Algorithm~\ref{alg:adpht_brief} gives the full policy.

\subsection{Reward Learner}
\label{sec:reward-module}
The first term $r$ of the score \eqref{eq:score}, the reward that a model solves the current task, is provided by the reward learner. Estimating it separately for every model wastes information because a difficult prompt tends to be difficult for several models, and a call to any model reveals something about that common difficulty. Pooling all calls into one predictor creates the opposite problem, because models have different strengths. The hybrid reward learner therefore combines a shared task component with a model-specific adjustment.

The reward feature $x_s^r$ of step $s$ is a high-dimensional prompt embedding, and the calls available to each model are limited in online routing. The shared component therefore works on the projected feature $z_s=Px_s^r$, where $P\in\R^{d_0\times d_r}$ is a fixed projection matrix with orthonormal rows and $d_0\ll d_r$, so the pooled block is estimated in a much smaller space. An unregularized fit would still be sensitive to directions that the executed calls sample poorly, so we estimate both components jointly under a ridge penalty, which limits the influence of those directions \citep{hastie2020ridge}.

Every executed call updates the shared sufficient statistics and the local sufficient statistics of the selected model. The fit returns a shared coefficient $\hat\beta_s\in\R^{d_0}$ and a coefficient $\hat\theta_{s,a}\in\R^{d_r}$ for each model $a$.
\begin{equation}
\hat r_s(a,x_s^r)=z_s^\top\hat\beta_s+{x_s^r}^\top\hat\theta_{s,a}.
\label{eq:hybrid-reward}
\end{equation}
The first term reads how difficult the task is for all models, and the second characterizes each model's specific performance on that task. This additive structure follows the hybrid linear upper-confidence-bound (LinUCB) algorithm \citep{li2010contextual}. The fixed projection reduces the shared block's dimension, motivated by random-projection methods for dimension reduction \citep{achlioptas2003database}. The detailed implementation is provided in \prettyref{app:repro}.

A point estimate alone would favor models already sampled often and could prematurely discard a capable but less-tested model. The learner therefore adds the uncertainty $v_s^r(a,x_s^r)$ of the same ridge fit, which is large when the current model--task pair is poorly covered by past calls:
\begin{equation}
U^r_s(a,x_s^r)=\clip_{[0,1]}\left\{\hat r_s(a,x_s^r)
+\alpha_r\sqrt{\max\{v^r_s(a,x_s^r),0\}}\right\} \,,
\label{eq:reward-ucb}
\end{equation}
where $\alpha_r>0$ is the reward-exploration scale, clipping keeps the score in $[0,1]$, and the inner $\max\{\cdot,0\}$ guards against a negative variance proxy produced by rounding. The selector uses $U^r_s$ rather than the point prediction, so fresh evidence can overturn an estimate built from few calls, leading to a balanced exploration--exploitation trade-off.

\subsection{Cost Learner}
\label{sec:cost-module}
\label{sec:kbit}

The term $c$ of \eqref{eq:score} is supplied by the cost learner. For a call to LLM $a$, the cost consists of two parts, an input charge and an output charge, with known per-token prices $\rho_a^{\rm in}$ and $\rho_a^{\rm out}$.
The input side of the charge is already known at admission (the prompt), so the quantity to forecast is the response length, for which the natural device is a point estimate. Such an estimate is unreliable, because a prompt does not determine a length. Even holding the model and the decoding configuration fixed, repeated calls on the same prompt trace out a heavy-tailed distribution of output lengths rather than one deterministic value \citep{wang2026prod}, so the estimate reports the center of a distribution whose spread is precisely what matters. The consequences of prediction error are asymmetric. Overprediction lowers a model's selection score and can discourage a retry that the budget could have afforded. Underprediction can instead favor a response whose realized charge crosses the budget and ends service for every later task. The selection score therefore needs a statement about the tail of the length distribution rather than only its center.

The $K$-bit mechanism takes the place of that point estimate. The resolution parameter $K$ determines the number of response-length intervals, $J=2^K$. Rather than predict precisely how many tokens a response will contain, the router first discretizes the range of possible lengths into $J$ intervals, then asks at each interval boundary whether the response will run past it, and reports the upper edge of an interval as its final prediction. This can lower the difficulty of the prediction, and the reason is easy to see intuitively. At $K=1$, for instance, the router predicts only whether a response will be short or long. That single bit can be easier to get right than an exact point estimate, and it also helps control the risk that a misprediction lets an overlong response exhaust the remaining budget. The price of the lost resolution is just as visible. A coarse grid limits how precisely the learner can locate the target response-length threshold. A larger $K$ supplies more thresholds, and every call updates all heads of its selected model. The trade-off comes from requiring confidence across more estimates, which increases the estimation allowance at a fixed sample size under the stated confidence calibration. The analysis following Algorithm~\ref{alg:adpht_brief} quantifies this trade-off between resolution and precision.

We call this coarsened signal the \emph{$K$-bit cost-shape channel}, and construct it as follows. In the formal model, the $J$ intervals are delimited by edges $0=e_0<e_1<\cdots<e_J\leq\ell_{\max}^{\rm out}$, where $\ell_{\max}^{\rm out}$ is the known maximum output length. These edges are chosen before the run and fixed during online decision processes. The edges can be constructed from the warm-start samples. For example, for $K=2$ and an equal-probability grid, the three interior edges are chosen to be the $25\%$, $50\%$, and $75\%$ quantiles of the historical response length distribution of warm-up calls or stored user datasets. In our implementation, we instead use warm-start quantiles to place $2^K-1$ intervals within a core range and reserve one interval for the modeled tail, where $e_J$ can be chosen and updated by the maximum observed length. This prevents the longest possible response length $\ell_{\max}^{\rm out}$ from influencing the core resolution. Thus $K=2$ gives three core intervals and one tail interval. The empirical top edge and the construction details are given in \prettyref{app:params}. At each threshold, we observe whether the response ran past the edge, an indicator we write $y^c_{s,j}=\1\{\ell_s^{\rm out}>e_j\}$. It then updates every threshold head of the selected model The formal analysis also allows other fixed, strictly increasing construction methods, such as equal-width partitions. For other grids, the extension in \prettyref{app:k-grid-extension} uses the distance to the next available edge. This term is no larger than the widest interval, $\max_{j\in[J]}(e_j-e_{j-1})$. Thus, the edge choice affects the size of the bound, but not its structure.

For candidate model $a$ and edge $e_j$, $j\in[J]$, the learner estimates the exceedance probability
\begin{align}
p^c_s(a,x_s,j)=\Pr\!\left(\tilde\ell_s^{\rm out}(a,x_s)>e_j\mid\calF_s\right).
\label{eq:survival-prob}
\end{align}

Here $\tilde\ell_s^{\rm out}(a,x_s)$ is the length model $a$ would return in state $x_s$, and the conditioning is on everything the router knows before the call. Note that since $e_J$ is an empirical upper bound in the implementation, actual response lengths may exceed it, so the probability in \eqref{eq:survival-prob} can be positive even for $j=J$.

For each model--edge pair $(a,j)$, a threshold head uses ridge regression fitted on past calls to model $a$ to predict the exceedance label $y^c_{s,j}$ from the cost feature $x_s^c(a)$, where $x_s^c(a)$ combines the projected prompt feature $z_s$ with pre-call information such as retry depth, input length, and the previous call's outcome and cost. With cost-exploration scale $\alpha_c>0$, the head reports the raw upper confidence bound
\begin{equation}
\begin{aligned}
\hat U^c_s(a,x_s^c(a),j)=
\clip_{[0,1]}\left\{\hat p_s(a,x_s^c(a),j)
+\alpha_c\sqrt{\max\{v_s^c(a,x_s^c(a)),0\}}\right\}.
\label{eq:cost-ucb}
\end{aligned}
\end{equation}
Here $\hat p_s(a,x_s^c(a),j)$ is the fitted exceedance probability, and $v_s^c(a,x_s^c(a))$ measures estimation uncertainty at the current cost feature. The uncertainty measure $v_s^c(a,x_s^c(a))$ is common to all $J$ heads of model $a$, because they use the same past calls, cost features, and ridge regularization and differ only in their exceedance labels. \prettyref{app:repro} gives the regression statistics and the expression for $v_s^c$ in detail.

True exceedance probabilities are nonincreasing in the threshold index $j$, since a response that exceeds $e_j$ also exceeds every shorter edge. But heads fitted separately need not respect that ordering in finite samples. The router restores it with a single sweep from short to long, replacing each bound by the smallest one seen so far as follows.
\begin{equation}
U^c_s(a,x_s^c(a),j)=
\min_{i\le j}\hat U^c_s(a,x_s^c(a),i).
\label{eq:monotone-cost-ucb}
\end{equation}
The reported bound $U_s^c(a,x_s^c(a),j)$ is therefore nonincreasing in the edge. We call this correction the \emph{monotone projection}, which restores coherence without adding information.

To obtain a response-length bound, fix a tail tolerance $\delta_c\in(0,1)$ specifying the maximum acceptable exceedance probability.\footnote{The tolerance can be tuned on validation data or chosen according to the user's risk preference. Holding the reported bounds fixed, decreasing $\delta_c$ will increase the selected edge and the resulting pre-call cost bound.} The \emph{first-safe-edge} rule selects the smallest edge whose reported bound $U^c_s(a,x_s^c(a),j)$ is at most $\delta_c$. If none qualifies, it falls back to the top edge $e_J$. The resulting edge length is
\begin{align}
\hat e_s^+(a,x_s^c(a))=
\min\left(\{e_j:j\in[J],\ U^c_s(a,x_s^c(a),j)\le\delta_c\}\cup \{e_J\}\right).
\label{eq:safe-edge}
\end{align}

Known token prices convert this length into the cost bound used in the selection score and retry rule,
\begin{align}
\hat c_s^+(a,x_s^c(a))=
\rho_a^{\rm in}\ell_s^{\rm in}(a)+\rho_a^{\rm out}\hat e_s^+(a,x_s^c(a)),
\label{eq:safecost}
\end{align}
where $\ell_s^{\rm in}(a)$ is the prompt length under model $a$'s own tokenizer, a function of $a$ because tokenizers differ enough across model families that one prompt yields a different token count, and hence a different input charge, on each candidate \citep{petrov2023tokenizer}.

\subsection{Dual-Pace Controller}
\label{sec:capacity-module}
The dual-pace controller sets the shadow price $\lambda_s$ in \eqref{eq:score} and the budget references used to evaluate each call. At decision step $s$, the remaining budget $R_s$ needs to support the $N_s$ tasks remaining in the stream, including the current task. Since the difficulty of future tasks is unknown, we use the equal share $\bar b_s=R_s/N_s$ as an initial spending reference. A strict per-task cap at this share can be too restrictive. If inexpensive tasks repeatedly leave part of their share unused, the share available to later tasks increases. A strict cap can therefore limit spending on earlier tasks even when budget is left unused at the end. Allowing unrestricted spending creates the opposite risk, because early tasks may exhaust the budget before later tasks are served.

\algname{} uses the dual-pace controller to manage this tradeoff. A pacing multiplier $\phi_s$ adjusts the spending reference for the current task based on recent average call costs. A shadow price $\lambda_s$ adjusts the cost penalty in model selection based on the cost of the latest call. The former responds to the recent spending pattern, while the latter responds directly to the latest call.

We define the task budget reference $b_s^{\rm task}$ and step budget reference $ b_s^{\rm step}$ as
\begin{subequations}
\label{eq:local-budget}
    \begin{align}
        b_s^{\rm task}
        &=\max\{\phi_s\bar b_s,\epsilon\},
        \label{eq:local-budget-task}\\
        b_s^{\rm step}
        &=\max\left\{
        \frac{b_s^{\rm task}}{H-h_s+1},\epsilon
        \right\},
        \label{eq:local-budget-step}
    \end{align}
\end{subequations}
where $H-h_s+1$ is the number of attempts remaining for the current task, including the current attempt. The task budget reference $b_s^{\rm task}$ provides a scale for the cost normalization, while the step budget reference $b_s^{\rm step}$ provides the spending reference for updating $\lambda_s$. The small constant $\epsilon>0$ keeps denominators positive.

After a call returns, the router observes its exact cost $c_s$ and updates an exponential moving average (EMA) of call costs,
\begin{equation}
    m_{s+1}=(1-\alpha_{\rm EMA})m_s+\alpha_{\rm EMA}c_s,
    \label{eq:cost-ema}
\end{equation}
where $\alpha_{\rm EMA}\in(0,1)$ is the weight on the latest charge. The initial average is obtained from the historical data.

The shadow price compares the latest charge with the step allowance, whereas the pacing multiplier compares the equal-share reference with the updated average cost. With learning rates $\eta_\lambda,\eta_\phi>0$, the two multiplicative updates are
\begin{subequations}
\begin{align}
\lambda_{s+1}
&=\clip_{[\lambda_{\min},\lambda_{\max}]}\!\left\{
\lambda_s
\exp\!\left(
\eta_\lambda
\clip_{[g_\lambda^{lb},g_\lambda^{ub}]}
\left\{\frac{c_s}{b_s^{\rm step}}-1\right\}
\right)
\right\},
\label{eq:lambda-eg}\\
\phi_{s+1}
&=\clip_{[\phi_{\min},\phi_{\max}]}\!\left\{
\phi_s
\exp\!\left(
\eta_\phi
\clip_{[g_\phi^{lb},g_\phi^{ub}]}
\left\{
\frac{\bar b_s}{\max\{m_{s+1},\epsilon\}}-1
\right\}
\right)
\right\},
\label{eq:phi-eg}
\end{align}
\end{subequations}
where the clipping bounds $g_\lambda^{lb},g_\lambda^{ub},g_\phi^{lb},g_\phi^{ub}$ prevent extreme updates, and the outer clips $[\lambda_{\min},\lambda_{\max}]$ and $[\phi_{\min},\phi_{\max}]$ keep the shadow price and pacing multiplier within reasonable ranges.

For each LLM call, if the realized cost $c_s$ exceeds the step allowance $b_s^{\rm step}$, the shadow price rises, lowering the score of expensive models in the next step. If the realized cost is below the allowance, the shadow price falls, leading to more aggressive spending. The pacing multiplier $\phi_s$ compares the equal-share reference with the moving average of recent costs. If the average cost is below the equal share, $\phi_s$ rises, allowing more budget for the current task. If the average cost exceeds the equal share, $\phi_s$ falls and thus tighten the budget. The use of a moving average makes pacing less sensitive to a single charge, while the shadow price responds to that charge directly. The dual-pace controller therefore adapts to both the realized spending path and the recent cost pattern, which helps avoid overspending and underspending adaptively.

\subsection{Model Selection and Retry Gate}
\label{sec:arm-selection}
\label{sec:retry-module}

With all modules above, the router computes the score for each model arm and selects the one with the highest score as follows:
\begin{subequations}
\label{eq:adpht-selection}
\begin{align}
Q_s(a,x_s)&=U^r_s(a,x_s^r)-\lambda_s\frac{\hat c_s^+(a,x_s^c(a))}{b_s^{\rm task}},
\label{eq:adpht-score}\\
a_s&=\argmax_{a\in\mathcal M}Q_s(a,x_s).
\label{eq:adpht-argmax}
\end{align}
\end{subequations}

The score \eqref{eq:adpht-score} aligns with the core design of score \eqref{eq:score}. The cost $\hat c_s^+$ is normalized by the task budget reference $b_s^{\rm task}$, which is dynamically adjusted by the remaining budget and the recent spending pattern. The shadow price $\lambda_s$ changes with the realized cost of the latest call, reflecting the current scarcity and value of the remaining budget. By multiplying the normalized cost by the shadow price, the cost can be translated into the same scale as the reward $U^r_s$, allowing a direct comparison. A model with higher possible reward and lower cost will have a higher score, and the models with higher rewards but higher costs may also be preferred. The score therefore balances the reward of solving the task with the cost of the call, and the model with the highest score is selected for step $s$. Ties are broken first by the optimistic solve probability and then by a mean-cost proxy, both recorded in \prettyref{app:repro}.

Equation~\eqref{eq:adpht-selection} determines which model to call. After a failed attempt, the router must decide whether to make another call or move to the next task. A further attempt can exploit the failed exchange, but it starts from a longer prompt and may also fail with higher cost. The retry gate therefore evaluates this choice before the next call using accumulated cost and remaining budget.

Suppose attempt $h_s<H$ fails and leaves $R_s-c_s>0$. The router records the exact charge, updates the dual-pace controller, and recomputes the task budget reference from the smaller post-call budget as $b_s^{\rm retry}=\max\{\phi_{s+1}\bar b_{s+1},\epsilon\}$ with $\bar b_{s+1}=(R_s-c_s)/N_s$, where $N_s$ is unchanged because the failed task is still unsolved. The exact spend on the current task through step $s$ is $\sum_{u\le s:\,t_u=t_s}G_uc_u$, where $G_u$ indicates whether step $u$ executes a call. Using this sum, the router computes a retry margin $\kappa_s$ that rises with retry depth and spending beyond the revised task budget reference as follows.
\begin{align}
\kappa_s=
\kappa_0+\kappa_h\frac{h_s-1}{H-1}
+\kappa_b\left(\frac{\sum_{u\le s:\,t_u=t_s}G_uc_u}{b_s^{\rm retry}}-1\right)_+,
\label{eq:retry-gate}
\end{align}
where $(u)_+:=\max\{u,0\}$. The base term $\kappa_0$ is the required retry margin, $\kappa_h$ penalizes deeper retries, and $\kappa_b$ penalizes overspending. This retry gate aligns with the LLM's practical behavior, because a retry after multiple failures is less likely to succeed, and the cost will be higher because the prompt is longer with the failed response appended \citep{iclr2026-llm-lost}. After computing $\kappa_s$ and rebuilding the next state $x_{s+1}$, the router reevaluates all candidates using the updated learners and controller. The task continues if and only if $\max_a Q_{s+1}(a,x_{s+1})\ge\kappa_s$. Algorithm~\ref{alg:adpht_brief} summarizes the overall execution of \algname{}.

\vspace{1em}
\begin{breakablealgorithm}
\caption{\algname{} policy.}
\label{alg:adpht_brief}
\begin{algorithmic}[1]
\Require Tasks $q_1,\ldots,q_T$, arms $\mathcal M$, budget $B$, retry limit $H$, resolution parameter $K$, tail tolerance $\delta_c$, and warm log $\mathcal D_{\rm warm}$ of calls executed before the stream.
\Ensure Solved-task count \textsc{Solved} and realized cost $C$.
\State Replay $\mathcal D_{\rm warm}$ to fix the edges and initialize both learners, $\lambda$, $\phi$, and $m$ and set $C\gets0$. \Comment{initialization}
\For{$t=1$ to $T$}
    \State Set $C^{\rm task}\gets0$ and $N\gets T-t+1$.
    \State Observe $q_t$ and build its first-attempt state $x$.
    \For{$h=1$ to $H$}
        \State \textbf{if} $C\ge B$ \textbf{then halt}.
        \State Compute $b^{\rm task}$ and $b^{\rm step}$ using~\eqref{eq:local-budget}.\Comment{pacing sets the local shares}
        \For{$a\in\mathcal M$}
            \State Compute $U^r(a)$ using~\eqref{eq:reward-ucb} and $\hat c^+(a)$ using~\eqref{eq:safecost}. \Comment{value and cost bound}
            \State $Q(a)\gets U^r(a)-\lambda\hat c^+(a)/b^{\rm task}$ as in~\eqref{eq:adpht-score}. \Comment{cost priced in value units}
        \EndFor
        \State Select $a^\star\in\argmax_{a\in\mathcal M}Q(a)$.
        \State \textbf{if} $Q(a^\star)<\kappa$ and $h\neq 1$ \textbf{then break}; otherwise execute the call. \Comment{retry gate}
        \State Observe $(r,c,\ell^{\rm out})$ and update both learners. \Comment{feedback and update}
        \State Set $C\gets C+c$ and $C^{\rm task}\gets C^{\rm task}+c$.
        \State Update $m$, $\lambda$, and $\phi$ in that order using~\eqref{eq:cost-ema}--\eqref{eq:phi-eg}.\Comment{price fast, pace slow}
        \State \textbf{if} $r=1$ \textbf{then} mark task $t$ solved and \textbf{break}.
        \State \textbf{if} $h=H$ or $C\ge B$ \textbf{then break}.
        \State Recompute $b^{\rm retry}$ and $\kappa$ using~\eqref{eq:retry-gate}, and append the failure and rebuild the next state.
    \EndFor
\EndFor
\end{algorithmic}
\end{breakablealgorithm}

For Algorithm \ref{alg:adpht_brief}, we capture the $K$-bit trade-off in the pre-call cost bound. A larger $K$ creates a finer grid and produces a cost bound with higher resolution, while requires more data and effort to update all $2^K$ threshold heads. We quantify this in the following theorem.

\begin{theorem*}[Informal]
\label{thm:ktradeoff_informal}
Given $\delta_c\in(0,1)$, the gap between the true conditional $(1-\delta_c)$ cost quantile $c_s^{\delta_c}(a)$ and the pre-call cost bound $\hat c_s^+(a, x_s^c(a))$ in \eqref{eq:safecost} satisfies the following order bound under the calibration, local tail-margin, core-coverage, and grid-spacing conditions in \prettyref{app:proofs}, with the remaining parameters fixed:
\begin{equation}
0\le\hat c_s^+(a, x_s^c(a))-c_s^{\delta_c}(a)
=O\!\left(
\underbrace{2^{-K}}_{\text{resolution error}}
+\underbrace{\sqrt{\frac{K}{n_{s,a}}}}_{\text{precision error}}
\right),
\label{eq:k-tradeoff}
\end{equation}
where $n_{s,a}$ is the number of all calls used to fit model $a$'s cost heads.
\end{theorem*}

A formal theorem statement and its proof are provided in \prettyref{app:proofs}. The resolution term comes from grid rounding and decreases as the grid becomes finer. The precision term is the allowance for simultaneous estimation across all $2^K$ threshold heads and increases with $K$ at a fixed sample size under the stated calibration. \prettyref{app:k-tightness} gives a matching construction establishing joint order tightness under the stated calibration w.r.t. \prettyref{eq:k-tradeoff}.

Holding reward estimates and controller values fixed, a larger estimated
cost bound $\hat c_s^+(a, x_s^c(a))$ can lower a model's score in
\eqref{eq:adpht-score}. The $K$-bit trade-off can therefore affect model
selection and retry decisions. \prettyref{app:k-proof} bounds the
one-step quantile-score loss, while Section~\ref{sec:ksweep} evaluates
tasks solved and \prettyref{app:calibration} checks empirical coverage,
providing practical guidance on choosing $K$.

We next examine how the value of tail protection varies with the number
of remaining tasks. Suppose we route $T$ identical tasks, with one call per task and total budget $B$, using two models with known distributions. The predictable model $P$ has deterministic cost $c$ and success probability $u$. The risky model $R$ has success probability $v>u$. It costs $\ell$ with probability $1-\varepsilon$ and $d\ge B$ with probability $\varepsilon$, where $\ell<c<d$. A response that reaches or crosses the budget is still completed and, if successful, counted before service stops.

It is easy to construct an instance in which every fixed mean-cost score prefers $R$. For example, take $(\ell,c,d,\varepsilon,u,v,B)=(1,4,9,0.2,0.75,1,7.5)$. Then $R$ has both a higher success probability and a lower mean cost. Thus, for every $\lambda\ge0$, the score \eqref{eq:score} prefers $R$.

However, the budget stopping rule changes this conclusion because a high-cost call to $R$ may end future service. With few tasks remaining, there is little future service to lose, so $R$ is better. As more tasks remain, avoiding termination protects enough future service to justify choosing $P$. With many tasks remaining, however, the extra cost of $P$ uses budget that could pay for several cheap calls to $R$, so $R$ becomes better again.

This example shows how the number of remaining tasks changes the optimal first choice. Even when $R$ has both a higher success probability and a lower mean cost, the optimal first choice moves from $R$ to $P$ and then back to $R$ as the number of remaining tasks increases. A fixed score whose predictions do not change with the number of remaining tasks cannot reproduce these changes. \algname{} uses a state-dependent cost penalty: the effective price $\lambda_s/b_s^{\rm task}$ on the cost bound in \eqref{eq:adpht-score} responds to realized spending and the remaining budget per unfinished task. The example yields the following structural insight.

\begin{remark}
\label{rmk:tail-reversal}
For a given budget, the value of tail protection can rise and fall as the number of remaining tasks increases. With few remaining tasks, there is little future service to protect. With many remaining tasks, the extra cost of tail protection displaces cheap risky calls. The optimal first choice can therefore be risky, then predictable, then risky again, which should be adaptively reflected in the model score.
\end{remark}

This analysis motivates a state-dependent cost penalty. \prettyref{app:tail-routing} gives the formal results and distinguishes the optimal price in the two-model setting from \algname{}’s adaptive rule; in particular, fixed initialization cannot reproduce both backlog reversals.

\section{Numerical Experiments}
\label{sec:experiments}
We conduct three experiments under the operating rules of Section~\ref{sec:model}. Experiment~1 (Section~\ref{sec:study-live}) compares the implemented policies on live \textsc{MATH500} \citep{math500} and \textsc{GPQA} \citep{rein2023gpqa} streams, with retries. It examines performance across budgets and sensitivity to the resolution parameter $K$, the retry gate, and pacing. Experiment~2 evaluates adapted configurations on a public benchmark (Section~\ref{sec:routerbench}), and Experiment~3 compares test-time allocation with a commercial router provided by OpenRouter (Section~\ref{sec:commercial}). We also conduct experiments in \prettyref{app:cost-head-exceedance} and \prettyref{app:dp-benchmark} to examine the $K$-bit mechanism through cost predictor ablations and assess how closely \algname{} approaches the dynamic programming optimum. All experiments run on a MacBook Air with an Apple M4 chip and 16GB of RAM.

We compare \algname{} against four implemented policies from the multi-LLM routing literature and a deployed commercial router. Budget-aware (BA) and Position-aware (PA) LinUCB \citep{poon2026multillm} are budget-aware contextual bandit routers. ParetoBandit \citep{tabernermiller2026paretobandit} is ported to the finite stream with its primal--dual reward--cost rule intact, and Greedy LinUCB always calls the arm with the highest UCB, standing in for a budget-unaware online router.

\subsection{Experiment 1: Routing and Retries Under a Binding Budget}
\label{sec:study-live}

We use a 20\%/80\% stratified split of each dataset into warm and test streams, with five shared-warm-start replicates per dataset. The warm stream is used to construct the warm-up data to initialize the predictors, which is excluded from budget accounting. And the budget is for the test stream only. A task may receive at most $H=4$ calls and leaves the stream after its first verifier-confirmed success, and the outcome is the number of distinct test tasks solved before service ends. The \textsc{MATH500} budget grid is $B\in\{0.025,0.05,0.075,0.125,0.30,0.45,0.60\}$ USD, to which \textsc{GPQA} adds $0.80$ and $1.00$. The tightest budgets cover only a small fraction of the stream, so the grid spans the scarcity regimes the model targets. Every policy faces the same five-model menu, task order, prompt constructors, verifier, local representation, seed-specific warm log, and cost accounting; the warm log is generated once by unconstrained Greedy LinUCB and replayed before each finite-budget run. \prettyref{app:eval-protocol} reports the details.

\subsubsection{Results}
\begin{figure}[!htbp]
\centering
\includegraphics[width=0.75\linewidth]{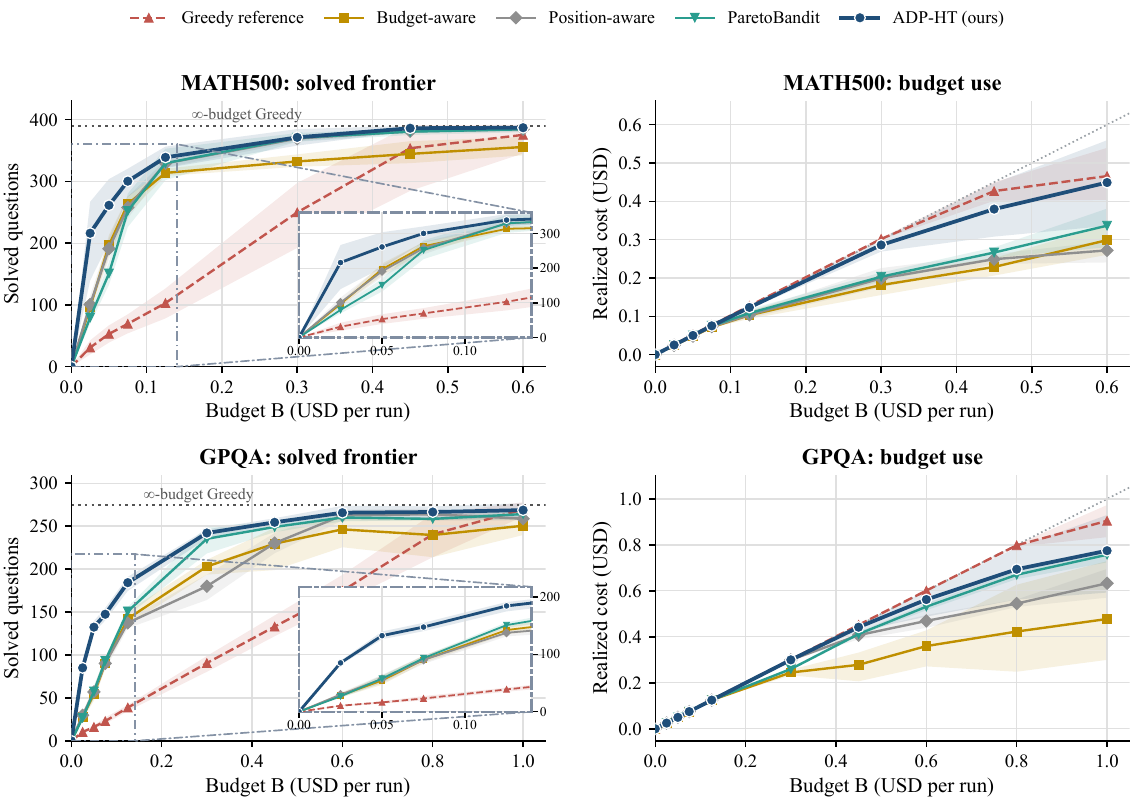}
\vspace{-4mm}
\caption{Performance on \textsc{MATH500} and \textsc{GPQA} streams.}
\label{fig:frontier}
\end{figure}

Figure~\ref{fig:frontier} reports the operating frontiers. \algname{}'s mean solved-count advantage over the budget-aware comparators is largest at tight budgets and narrows as the budget increases. At $B=0.025$ it solves $216.2$ \textsc{MATH500} problems and $85.0$ \textsc{GPQA} questions against $100.8$ and $29.8$ for Position-aware LinUCB, the strongest implemented comparator in both datasets, or more than twice as many in each. At the largest budget of each dataset, the advantage over the best of BA, PA, and ParetoBandit is $2.6$ and $4.6$ tasks. Budget-unaware Greedy, far behind at tight budgets, closes to within twelve tasks on \textsc{MATH500} and has a similar mean solved count on \textsc{GPQA}.

Converting the budget into completions is not the same objective as spending little. At $B=0.30$ on \textsc{GPQA}, ParetoBandit solves $235.2$ questions for about $\$0.26$, whereas \algname{} solves $242.0$ for about $\$0.30$. \algname{} nevertheless has the higher solved-count mean at every tested budget, with grid-averaged advantages of $44.8$ tasks on \textsc{MATH500} and $27.9$ on \textsc{GPQA}. Because the same warm replay re-enters the analysis at every budget of its seed, budget cells are not independent replicates. A two-level paired bootstrap of that difference yields $95\%$ percentile intervals entirely above zero in all six dataset--baseline comparisons. \prettyref{app:tables} reports its construction and the cell-level means behind Figure~\ref{fig:frontier}.

The realized paths help describe how \algname{} uses the budget. It uses $93.4\%$ of the nominal budget on average across the \textsc{MATH500} grid, against $75.7$--$81.5\%$ for BA, PA, and ParetoBandit. This greater utilization is consistent with pacing releasing unused budget, although the comparison does not isolate its contribution. As the budget expands, the policy shifts calls from inexpensive models toward more capable ones. Calls per task on \textsc{MATH500} first rise as second attempts become affordable and then fall when stronger first calls enter, whereas on \textsc{GPQA} retries remain common because first attempts fail more often. The remaining unsolved tasks account for a disproportionate share of spending. Here we denote profile as the complete attempt history of one task actually started in one seed's run, and tasks never reached by the router are excluded. At the largest \textsc{MATH500} budget, unsolved profiles make up $3.25\%$ of these recorded profiles but absorb $20.2\%$ of spending. On \textsc{GPQA}, their share of spending is $53.9\%$. These figures show why neither low spending nor budget exhaustion alone measures allocation quality. \prettyref{app:tables} records the underlying utilization, profile counts, and per-task costs.

\subsubsection{Choosing the Resolution Parameter \texorpdfstring{$K$}{K}.}
\label{sec:ksweep}

\begin{figure}[!htbp]
\centering
\includegraphics[width=.85\linewidth]{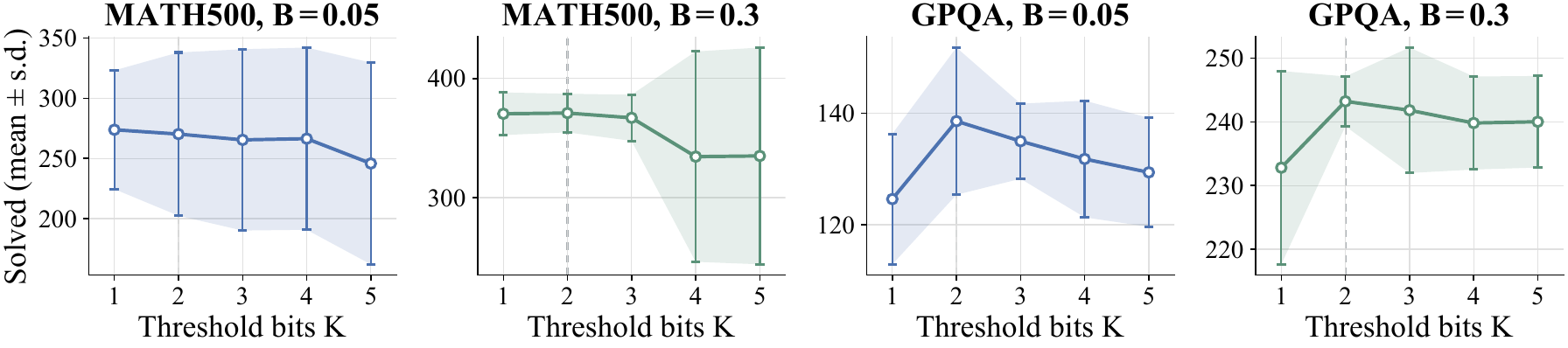}
\vspace{-4mm}
\caption{$K$-sweep in Experiment~1.}
\label{fig:kablation}
\end{figure}

Theorem~\ref{thm:ktradeoff_informal} identifies the trade-off between resolution and precision but does not select an operating point. On live traffic the resolution sweep favors a coarse mesh. The sweep shows no clear gain beyond the compact $K=2$ used in Figure~\ref{fig:frontier}, and empirical coverage declines at the finest meshes. Two groups of new runs examine these outcomes. A sweep reruns the full policy at $K\in\{1,\ldots,5\}$ in four dataset--budget cells, changing only the resolution parameter $K$, and a call-level audit compares the deployed bound with the charges it admitted. In the sweep (Figure~\ref{fig:kablation}), across-seed dispersion grows several-fold on \textsc{MATH500} at the two finest meshes; a two-level bootstrap pooled over the four cells separates $K=2$ from $K=4$ and $K=5$, but not from $K=1$ or $K=3$. The audit checks the coverage of the deployed cost bounds. These runs reuse warm data and tune the confidence scales, so the audit does not verify the theorem's calibration assumptions. The observed fraction of calls satisfying $c_s\le\hat c_s^+$ declines monotonically in $K$ on both datasets, and on \textsc{GPQA} it falls below the operating target $1-\delta_c$ at $K\ge4$. Provider variation and differences in routing history also affect these results. They support using a coarse mesh at the reported sample sizes, but do not isolate the estimation term or establish an optimal $K$. \prettyref{app:kablation} and~\prettyref{app:calibration} report the sweep design, cell-level counts, mean differences, and coverage audit.

\subsubsection{Ablation Study.}
\label{sec:ablation}

The ablation removes the retry gate and the pacing update, separately and together, at $B=0.05$ and $B=0.30$ with three seed indices per cell. The largest observed loss from removing the retry gate occurs at the tighter budget. At $B=0.05$ the no-gate variant solves about a third fewer \textsc{MATH500} tasks and under half as many \textsc{GPQA} tasks as the full policy, whereas at $B=0.30$ the gaps narrow to $7.4$ and $23.0$ tasks. Freezing the pacing multiplier at its initial value costs little on \textsc{MATH500} but $12.7$ tasks on \textsc{GPQA} at the tight budget. The no-gate gap is smaller at the larger tested budget, while the loss associated with freezing pacing varies by dataset and budget. These comparisons describe sensitivity to component removal. Table~\ref{tab:app-ablation} reports all cells and the seed protocol.

\newsavebox{\ablationtablebox}
\begin{table}[!htbp]
\centering
{\caption{Ablation study solved counts. Budgets are in USD.}
\label{tab:app-ablation}
\footnotesize
\sbox{\ablationtablebox}{%
\begin{tabular}{llcccc}
\toprule
Dataset & Budget & \algname{} & Frozen-$\phi$ & No-gate & Frozen-$\phi$+No-gate \\
\midrule
\textsc{MATH500} & 0.05 & 291.3 ($\pm28.9$) & 286.3 ($\pm2.1$) & 192.3 ($\pm16.3$) & 198.7 ($\pm7.6$) \\
 & 0.30 & 371.7 ($\pm15.9$) & 365.7 ($\pm6.7$) & 364.3 ($\pm21.0$) & 366.7 ($\pm6.5$) \\
\midrule
\textsc{GPQA} & 0.05 & 129.7 ($\pm7.6$) & 117.0 ($\pm1.7$) & 63.0 ($\pm3.5$) & 53.7 ($\pm10.5$) \\
 & 0.30 & 246.0 ($\pm1.0$) & 239.3 ($\pm8.0$) & 223.0 ($\pm3.0$) & 239.0 ($\pm12.5$) \\
\bottomrule
\end{tabular}
}%
\begin{minipage}{\wd\ablationtablebox}
\usebox{\ablationtablebox}
\par\vspace{1mm}
\scriptsize Mean over three seeds, $\pm$~one standard deviation.
\end{minipage}}
\end{table}

\subsection{Experiment 2: Comparison on a Public Benchmark}
\label{sec:routerbench}
\begin{figure}[!htbp]
\centering
\includegraphics[width=0.85\linewidth]{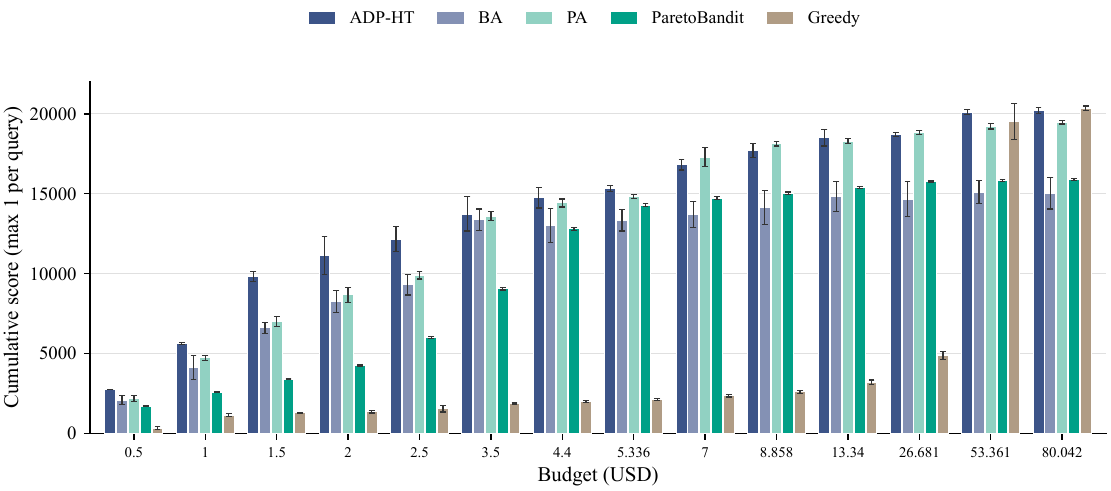}
\vspace{-4mm}
\caption{Comparison on \textsc{RouterBench}.}
\label{fig:routerbench-comparison}
\end{figure}

We also compare on \textsc{RouterBench} \citep{hu2024routerbench}, a public benchmark of 36,497 queries drawn from eight datasets and answered by eleven models under a single-call limit, with a standardized test stream, model costs, and a stored score in $[0,1]$ per query--model pair. \algname{} runs against BA, PA, ParetoBandit, and budget-unaware Greedy LinUCB over five seeds, each on a stratified $20\%/80\%$ split, across fourteen budgets spanning $\$0.50$ to $\$80.04$. The outcome is the five-seed mean \emph{cumulative score} of all tasks. Figure~\ref{fig:routerbench-comparison} plots it with the budgets ordered as categories, because the grid spans more than two orders of magnitude. \algname{} leads all eight budgets through $B=\$5.336$, reaching $9{,}827$ at $B=\$1.50$ against $6{,}982$ for the nearest comparator and $1{,}283$ for Greedy. It exceeds BA and ParetoBandit in every cell and leads all five protocols in ten of fourteen, with a largest deficit of $2.8\%$ in the four remaining cells. \prettyref{app:routerbench-protocol} gives the configurations, the warm information available to each method, and the cell-level means.

We also conduct a cost-head ablation on \textsc{RouterBench} within a common reward learner and budget controller. \textbf{K-Bit} is the $K$-bit head implemented in \algname{}. \textbf{Continuous Mean} uses the corresponding continuous output-length mean predictor. \textbf{Mean-Adjusted} adds a continuous-length confidence correction to Continuous Mean. K-Bit and Mean-Adjusted share the same tail-tolerance setting $\delta_c$.

\begin{table}[!htbp]
\centering
\caption{Cost-head comparison on \textsc{RouterBench}.}
\label{tab:routerbench-cost-head-main}
\footnotesize
\setlength{\tabcolsep}{4pt}
\begin{tabular}{lrrr}
\toprule
Budget (USD) & K-Bit & Continuous Mean & Mean-Adjusted \\
\midrule
5.00 & \textbf{15,989.4 (770.9)} & 14,774.2 (768.9) & 15,502.3 (416.7) \\
8.86 & \textbf{18,397.6 (462.3)} & 16,923.7 (1,589.2) & 17,419.5 (1,048.8) \\
10.00 & \textbf{18,673.3 (376.0)} & 17,248.3 (1,359.2) & 17,446.2 (1,199.9) \\
13.34 & \textbf{18,931.1 (515.2)} & 17,851.0 (1,330.5) & 17,762.9 (1,310.1) \\
20.00 & \textbf{18,996.2 (532.9)} & 18,771.0 (883.9) & 18,668.4 (992.8) \\
40.02 & 19,888.3 (315.2) & \textbf{20,072.6 (151.7)} & 20,054.0 (188.7) \\
80.04 & 20,272.6 (187.2) & 20,263.1 (195.8) & \textbf{20,283.6 (184.1)} \\
\bottomrule
\end{tabular}
\par\smallskip
\begin{minipage}{0.98\linewidth}\footnotesize
Entries are five-seed means with sample standard deviations. Bold indicates the highest mean at each budget.
\end{minipage}
\end{table}

Table~\ref{tab:routerbench-cost-head-main} compares seven representative
budgets over five runs with different seeds. K-Bit has the highest mean score at tight budgets, which shows the benefit of a discrete tail-safe predictor. And as the budget grows, all the three heads perform similarly, which is intuitive because the risk of exhausting the budget too early is lower. \prettyref{app:cost-head-exceedance} reports the grid diagnostic and all paired intervals.

\subsection{Experiment 3: Comparison with a Deployed Commercial Router}
\label{sec:commercial}
Experiment~3 compares \algname{} with the OpenRouter AutoRouter service \citep{openrouter2026autorouter}, using its recorded default and cost-saving configurations. We use \textsc{LiveCodeBench} \citep{jain2025livecodebench} release\_v6, with 844 test problems after a fixed-seed split reserves 211 for \algname{}'s warm start. Both routers share the frozen 39-model pool, problem order, one-call limit, and process-isolated execution grader, and use API-reported cost accounting.

\begin{figure}[!htbp]
\centering
\includegraphics[width=0.75\linewidth]{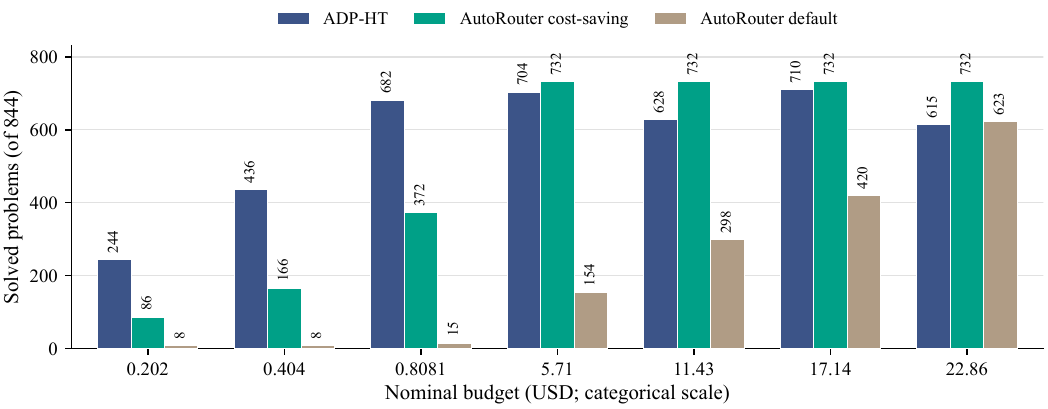}
\vspace{-4mm}
\caption{Commercial-router comparison on the common 844-problem \textsc{LiveCodeBench} stream.}
\label{fig:commercial-bars}
\end{figure}

Against a commercial router, \algname{}'s advantage is concentrated where the budget binds, and there it is large.  At the three lowest budgets, $\$0.202$, $\$0.404$, and $\$0.8081$, it essentially exhausts each one and solves $244$, $436$, and $682$ problems, against $86$, $166$, and $372$ for cost-saving AutoRouter and at most $15$ for the default setting (Figure~\ref{fig:commercial-bars}). In the two larger of those cells, it also spends less than cost-saving AutoRouter. Its 682-problem point exceeds default AutoRouter's complete path of 623 problems at $\$22.86$ in solved count and uses less test-time spend. Each AutoRouter curve is one complete pass truncated at the first budget crossing, so it cannot fall as $B$ grows, whereas every \algname{} point is a separate live run whose learning path and provider responses differ from cell to cell even at temperature zero. With only one run per budget, this response and execution variability is not averaged out and influenced by the LLM service provider, so \algname{}'s counts may not rise monotonically.

At larger budgets, call failures complicate the comparison. Cost-saving AutoRouter issues all 844 calls for $\$1.62$ and solves 732 problems, more than \algname{} reaches at any of the four larger budgets and at lower realized spend. All four of those \algname{} paths encounter provider-credit or relay-transport failures, reported in Table~\ref{tab:app-commercial}. We retain these calls as unsuccessful. The failures depress the recorded counts and prevent a clean comparison of routing performance at the larger budgets. This single-split study provides a snapshot of the two routers' performance on a shared stream, and highlights the importance of an adaptive and budget-aware routing policy.  \algname{} learns from its warm-start calls and keeps updating during service. The earlier AutoRouter service was described as using a pretrained router \citep{notdiamond2026router}, whose training-data size was not disclosed but can be expected to be large. This also uncovers another advantage of \algname{}, which can learn online and dynamically even after deployment, whereas a pretrained router cannot.
\section{Conclusion}
\label{sec:conclusion}We study LLM routing as an online allocation problem in which a call's cost is learned only after the budget has been committed. Under a halt-after-crossing budget, an admission decision affects both the current task and every later task the remaining budget could have served, so controlling spending in expectation does not control that exposure. We formulate the problem as $K$-bit Pathwise Contextual Bandits with Knapsacks. Exact charges accumulate in the ledger, while the cost learner observes only which of $2^K$ response-length intervals a call landed in. \algname{} turns those labels into a one-sided pre-call cost bound and prices it by the current shadow value of capacity, so one score governs both model choice and retries. The analysis gives a conditional $K$-bit trade-off for the implemented quantile mesh. More bits reduce core discretization but increase the allowance for simultaneous estimation, with an additional term for edge-placement error. A matching construction establishes the tightness of the two $K$-dependent terms under the stated confidence calibration. This bound motivates choosing resolution to match the available sample size. Under the protocols we report, the gains concentrate where the budget determines which tasks are reached. At the tightest tested budgets \algname{} completes more than twice as many \textsc{MATH500} and \textsc{GPQA} tasks as the best implemented comparator, leads ten of fourteen \textsc{RouterBench} budgets, and outperforms a deployed commercial router at the three lowest test budgets.

\emph{Future work.}~Several directions remain open. First, extending pathwise budget control to open-ended work such as writing, analysis, and code review would require modeling the uncertainty of a learned judge and its rubric-based evaluations. The resolution--estimation trade-off could then appear in the grading mesh as well as the length mesh, which makes that setting a natural test of how far the mechanism travels. Second, a joint treatment of selection, pricing, pacing, and retries would bring the dual path inside the guarantee and deliver an end-to-end pathwise statement. Lower bounds over routing policies would clarify how much additional service can be gained from finer cost information. Third, sharper response-length prediction is an immediately practical lever. Any predictor that localizes the conditional tail more tightly shrinks the estimation radius at a fixed resolution, so the same mesh admits a tighter bound, and the mechanism composes with advances made elsewhere. Fourth, an adaptive mesh that re-places its edges and reset $K$ as the workload drifts,  would make the resolution parameter $K$ and edge positions a control component that the operator can tune during service rather than a constant fixed at warm start.

\bibliographystyle{alpha}
\bibliography{ref}

\newpage
\appendix
\renewcommand{\thesection}{EC.\arabic{section}}

\addtocontents{toc}{\protect\setcounter{tocdepth}{1}}
\addtocontents{toc}{\protect\setlength{\sectiontocnumwidth}{3.5em}}

\section{Proof and Calibration Details}
\label{app:proofs}
We prove \eqref{eq:k-tradeoff} and derive its model-selection implication. We then give statistical conditions for the estimation bound, show that both error terms can attain the stated orders, and extend the result to other grids.

\subsection{Formal Cost-Bound Theorem}
\label{app:k-conditions}
Fix $(s,a)$, suppress the feature arguments, and write $p_s(a;\ell)=\Pr(\tilde\ell_s^{\rm out}(a)>\ell\mid\calF_s)$. For tail tolerance $\delta_c\in(0,1)$, define
\begin{equation}
q_s(a)=\inf\{\ell\ge0\mid p_s(a;\ell)\le\delta_c\},\qquad
c_s^{\delta_c}(a)=\rho_a^{\rm in}\ell_s^{\rm in}(a)+\rho_a^{\rm out}q_s(a).
\label{eq:true-safe-cost}
\end{equation}
Let $\hat c_s^+(a)$ denote \eqref{eq:safecost}. We study the excess $\hat c_s^+(a)-c_s^{\delta_c}(a)$. If $q_s(a)$ is finite, right-continuity ensures $p_s(a;q_s(a))\le\delta_c$.

At grid edges, $p_s(a;e_j)=p_s^c(a,x_s,j)$ by \eqref{eq:survival-prob}. Let $U_s^c(a,j)$ and $\hat e_s^+(a)$ be the bound and selected length from \eqref{eq:monotone-cost-ucb}--\eqref{eq:safe-edge}.
\label{app:k-grid}

For integer $K\ge1$, the frozen mesh $0=e_0<\cdots<e_J$ has $J=2^K$ intervals, with $J_c=J-1$ core intervals and one tail interval. A nondecreasing reference curve $\mathsf Q:[0,1]\to[0,\infty)$ assigns core-edge reference locations $\mathsf Q(j/J_c)$ and need not describe the current length distribution. Require $\eta\ge0$, $L_Q>0$, and
\begin{equation}
\begin{aligned}
|e_j-\mathsf Q(j/J_c)|&\le\eta,\quad j=0,\ldots,J_c,\\
|\mathsf Q(v)-\mathsf Q(u)|&\le L_Q|v-u|,\quad u,v\in[0,1].
\end{aligned}
\label{eq:k-quantile-regularity}
\end{equation}
Here $\eta$ bounds finalized edge error in tokens from warm-data estimation, integer lengths, and repeated-edge adjustments, without an assumed sampling rate.

Define the maximum actual core spacing $h_K=\max_{1\le j\le J_c}(e_j-e_{j-1})$ and reference spacing $\Delta_{Q,K}=\max_{1\le j\le J_c}[\mathsf Q(j/J_c)-\mathsf Q((j-1)/J_c)]$. Equation~\eqref{eq:k-quantile-regularity} gives
\[
h_K\le\Delta_{Q,K}+2\eta\le\frac{L_Q}{2^K-1}+2\eta.
\]
For $h_K\le C_h2^{-K}$ with $C_h$ independent of $K,n_{s,a}$, sufficient placement conditions are
\begin{equation}
\begin{aligned}
\eta\le\Delta_{Q,K}&\quad\Longrightarrow\quad C_h=6L_Q,\\
\eta\le D_\eta2^{-K}&\quad\Longrightarrow\quad C_h=2L_Q+2D_\eta,
\end{aligned}
\label{eq:k-spacing-constants}
\end{equation}
where $L_Q$ and $D_\eta\ge0$ are fixed across grids. These follow from $1/(2^K-1)\le2\,2^{-K}$. Exact placement permits $C_h=2L_Q$. The full bound retains $\eta$ without either spacing condition.

Let $r_{s,K}\ge0$ bound the excess of every reported probability bound. All threshold heads share $n_{s,a}=|\{u<s:G_u=1,\ a_u=a\}|\ge1$ warm ($u\le0$) and earlier online fitting calls in \eqref{eq:cost-design}, excluding separate grid-placement pilot data under \prettyref{app:k-confidence}. Require
\begin{gather}
0\le U_s^c(a,j)-p_s(a;e_j)\le r_{s,K},\qquad j\in[J],
\label{eq:k-confidence-excess}\\
r_{s,K}=C_c\sqrt{\frac{A+K\log2}{n_{s,a}}}.
\label{eq:k-confidence-scale}
\end{gather}
The constants $C_c,A>0$ are independent of $K$. Here $A$ collects feature, fitting-horizon, and confidence parameters, while $K\log2=\log(2^K)$ accounts for simultaneous confidence across thresholds. \prettyref{app:k-confidence} gives a calibration.

To translate probability error into a length allowance, let $u$ denote tokens added to $q_s(a)$. In addition to the grid and estimation conditions above, assume that there are constants $\nu,u_0>0$ satisfying
\begin{gather}
p_s(a;q_s(a)+u)\le\delta_c-\nu u,\qquad 0\le u\le u_0,
\label{eq:tail-margin}\\
\frac{r_{s,K}}{\nu}\le u_0,\qquad
q_s(a)+\frac{r_{s,K}}{\nu}\le e_{J_c}.
\label{eq:k-core-conditions}
\end{gather}
Here $\nu$ bounds the probability margin per added token from below over the range $u_0$. The last two inequalities keep the length allowance within this range and the core. The bound below holds at every history satisfying the grid, estimation, and margin conditions. It retains $\eta$ and $A$. The simplified rate also requires core spacing to decrease as $2^{-K}$ and $A$ to remain fixed.

The local margin condition \eqref{eq:tail-margin} in Theorem~\ref{thm:ktradeoff} does not require a continuous response-length distribution. It holds, for example, if $p_s(a;\cdot)$ is absolutely continuous on $[q_s(a),q_s(a)+u_0]$ and $-\partial_\ell p_s(a;\ell)\ge\nu$ almost everywhere on that interval. Since $p_s(a;q_s(a))\le\delta_c$, integration gives $p_s(a;q_s(a)+u)\le\delta_c-\nu u$. The length allowance $r_{s,K}/\nu$ must still satisfy \eqref{eq:k-core-conditions}.

\begin{theorem}[$K$-bit cost-bound trade-off]
\label{thm:ktradeoff}
At a pre-call pair $(s,a)$, use the grid and notation above with $K,n_{s,a}\ge1$, finite $q_s(a)$, input charge known at $\calF_s$, and $\rho_a^{\rm out}>0$. Construct the bounds and selected edge by \eqref{eq:monotone-cost-ucb}--\eqref{eq:safecost}, replacing only the fallback $\ell_{\max}^{\rm out}$ in \eqref{eq:safe-edge} by an empirical top edge $e_J$ when applicable. Assume the following conditions hold at this history.
\begin{enumerate}[label=(\roman*),leftmargin=*]
\item The finalized core grid satisfies \eqref{eq:k-quantile-regularity}.
\item All reported bounds satisfy \eqref{eq:k-confidence-excess}--\eqref{eq:k-confidence-scale}.
\item The conditional tail and core coverage satisfy \eqref{eq:tail-margin}--\eqref{eq:k-core-conditions}.
\end{enumerate}
A core edge passes the first-safe-edge test, and
\begin{equation}
\begin{aligned}
0\le\hat c_s^+(a)-c_s^{\delta_c}(a)
&\le\rho_a^{\rm out}\left(h_K+\frac{r_{s,K}}{\nu}\right)\\
&\le\rho_a^{\rm out}\left[
\frac{L_Q}{2^K-1}+2\eta
+\frac{C_c}{\nu}\sqrt{\frac{A+K\log2}{n_{s,a}}}
\right].
\end{aligned}
\label{eq:k-tradeoff-explicit}
\end{equation}
For the potential call cost $\tilde c_s(a)=\rho_a^{\rm in}\ell_s^{\rm in}(a)+\rho_a^{\rm out}\tilde\ell_s^{\rm out}(a)$, the same history satisfies
\begin{equation}
\Pr\!\left(\tilde c_s(a)>\hat c_s^+(a)\mid\calF_s\right)\le\delta_c.
\label{eq:k-conditional-coverage}
\end{equation}
If, in addition, $h_K\le C_h2^{-K}$ and $C_h,C_c,A,\nu$, and $\rho_a^{\rm out}$ are fixed independently of $K$ and $n_{s,a}$, then
\begin{equation}
0\le\hat c_s^+(a)-c_s^{\delta_c}(a)
\le\rho_a^{\rm out}\left[
C_h2^{-K}+\frac{C_c\sqrt{A+\log2}}{\nu}\sqrt{\frac{K}{n_{s,a}}}
\right].
\label{eq:k-tradeoff-two-term}
\end{equation}
Thus \eqref{eq:k-tradeoff} holds uniformly over the pairs $(K,n_{s,a})$ satisfying these conditions, with a constant independent of $K$ and $n_{s,a}$.
\end{theorem}

\subsection{Proof and Model Selection}
\label{app:k-proof}
\noindent\emph{Proof of Theorem \ref{thm:ktradeoff}.}
Write $q=q_s(a)$ and $r=r_{s,K}$. For each finalized core interval, \eqref{eq:k-quantile-regularity} gives
\begin{equation}
e_j-e_{j-1}
\le \mathsf Q(j/J_c)-\mathsf Q((j-1)/J_c)+2\eta
\le \frac{L_Q}{J_c}+2\eta.
\label{eq:app-core-spacing}
\end{equation}
Set $t=q+r/\nu$ and take the first core edge $e_j\ge t$, which exists by \eqref{eq:k-core-conditions}. Then $e_j-t\le e_j-e_{j-1}$, also for $j=1$ since $e_0=0\le t$. Monotonicity and \eqref{eq:tail-margin} give $p_s(a;e_j)\le p_s(a;t)\le\delta_c-r$ and $U_s^c(a,j)\le p_s(a;e_j)+r\le\delta_c$. Thus $e_j$ qualifies and coverage yields $q\le\hat e_s^+(a)\le e_j$, so
\[
0\le\hat e_s^+(a)-q
\le(e_j-t)+(t-q)
\le h_K+\frac{r}{\nu}
\le\frac{L_Q}{2^K-1}+2\eta+\frac{r}{\nu}.
\]
The input charge cancels. Multiplying by $\rho_a^{\rm out}$ and substituting \eqref{eq:k-confidence-scale} proves \eqref{eq:k-tradeoff-explicit}.
For \eqref{eq:k-tradeoff-two-term}, substitute $h_K\le C_h2^{-K}$ and $A+K\log2\le(A+\log2)K$ in \eqref{eq:k-tradeoff-explicit}. Equation~\eqref{eq:k-spacing-constants} supplies sufficient $C_h$. The implicit constant can be $\rho_a^{\rm out}\max\{C_h,C_c\sqrt{A+\log2}/\nu\}$, independent of $K,n_{s,a}$ under the fixed-parameter conditions. The calibration below requires fixed $N$, feature and regularization parameters, and confidence level. If $N$ grows with $n_{s,a}$, retain $A$ as in \eqref{eq:k-tradeoff-explicit}.
Cost exceeds its bound exactly when output length exceeds the selected edge, giving \eqref{eq:k-conditional-coverage}. A qualifying core edge avoids any need for a support bound at $e_J$.\hfill$\square$

At an executed step with controller values $\lambda_s\ge0$ and $b_s^{\rm task}>0$, define
\begin{align}
\mu_s(a)&=\E[\tilde r_s(a,x_s)\mid\calF_s]
=\Pr(\tilde r_s(a,x_s)=1\mid\calF_s),
\label{eq:true-solve-prob}\\
g_s(a)&=\mu_s(a)-\frac{\lambda_s}{b_s^{\rm task}}c_s^{\delta_c}(a),
\qquad a_s^\star\in\argmax_{a\in\mathcal M}g_s(a).
\notag
\end{align}
For $a_s$ selected by \eqref{eq:adpht-argmax}, let $E_{s,K}(a)$ denote \eqref{eq:k-tradeoff-explicit}. If the theorem holds for $a_s,a_s^\star$ and both reward estimates satisfy $|\hat r_s(a,x_s^r)-\mu_s(a)|\le\alpha_r\sqrt{\max\{v_s^r(a,x_s^r),0\}}$, then
\begin{equation}
\begin{aligned}
0&\le g_s(a_s^\star)-g_s(a_s)\\
&\le 2\alpha_r\sqrt{\max\{v_s^r(a_s,x_s^r),0\}}
+\frac{\lambda_s}{b_s^{\rm task}}E_{s,K}(a_s^\star).
\end{aligned}
\label{eq:k-choice-loss}
\end{equation}

Put $\tau=\lambda_s/b_s^{\rm task}\ge0$, $d_a=U_s^r(a,x_s^r)-\mu_s(a)$, and $\epsilon_a=\hat c_s^+(a)-c_s^{\delta_c}(a)$. For both arms, clipping and reward confidence give $0\le d_a\le2\alpha_r\sqrt{\max\{v_s^r(a,x_s^r),0\}}$, and the theorem gives $0\le\epsilon_a\le E_{s,K}(a)$. Maximization of $Q_s(a,x_s)=g_s(a)+d_a-\tau\epsilon_a$ yields
\[
\begin{aligned}
0\le g_s(a_s^\star)-g_s(a_s)
&\le d_{a_s}-d_{a_s^\star}+\tau(\epsilon_{a_s^\star}-\epsilon_{a_s})\\
&\le2\alpha_r\sqrt{\max\{v_s^r(a_s,x_s^r),0\}}+\tau E_{s,K}(a_s^\star).
\end{aligned}
\]
This proves \eqref{eq:k-choice-loss}.\hfill$\square$

\subsection{Statistical Conditions for the Estimation Bound}
\label{app:k-confidence}
Fix a deterministic fitting-call bound $N$ per model and $\delta_{\rm stat}\in(0,1)$. The following bounds hold on one event of probability at least $1-\delta_{\rm stat}$ over fitting data.

Fix a deterministic mesh before collecting fitting data, or construct it from separate pilot data and condition all assumptions on those data. Exclude pilot data from fits. At every warm or online fitting call, model choice and cost feature must be known before the response, but may depend on earlier responses. For each $(a,j)$, a fixed coefficient $\theta^c_{a,j}\in\mathbb R^{d_c}$ applies to fitting and prediction, with $p_s(a;e_j)=x_s^c(a)^\top\theta^c_{a,j}$, $\|x_s^c(a)\|\le L_c$, and $\|\theta^c_{a,j}\|\le S_c$.
Extend $\calF_u$ to warm calls, including separate pilot data in the initial information. For every fitting call $u$ to model $a_u$, threshold $j$, and $t\in\mathbb R$, require
\[
\begin{aligned}
\E[y^c_{u,j}\mid\calF_u]&=x_u^c(a_u)^\top\theta^c_{a_u,j},\\
\E\!\left[\exp\!\left\{t\bigl(y^c_{u,j}-x_u^c(a_u)^\top\theta^c_{a_u,j}\bigr)\right\}\mid\calF_u\right]
&\le\exp(t^2R_c^2/2).
\end{aligned}
\]
All conditions include warm observations. Binary labels permit $R_c=1/2$. Let $\gamma_c>0$ denote ridge regularization, $V^c_{s,a}$ the design in \eqref{eq:cost-design}, and $I_{d_c}$ the identity.

Apply the self-normalized ridge bound of \citet{abbasi2011improved} to each head with failure probability $\delta_{\rm stat}/(M2^K)$. The bound is simultaneous over fitting times and therefore permits adaptive model choices. The determinant bound
\[
\frac{\det V^c_{s,a}}{\det(\gamma_c I_{d_c})}
\le\left(1+\frac{n_{s,a}L_c^2}{\gamma_cd_c}\right)^{d_c}
\le\left(1+\frac{NL_c^2}{\gamma_cd_c}\right)^{d_c}
\]
and a union bound over $M2^K$ heads, requiring no independence between them, give the simultaneous prediction bounds
\begin{gather}
\beta_K=
R_c\sqrt{d_c\log\!\left(1+\frac{NL_c^2}{\gamma_cd_c}\right)
+2\log\!\left(\frac{M2^K}{\delta_{\rm stat}}\right)}
+\sqrt{\gamma_c}S_c,
\label{eq:exact-cost-radius}\\
\left|\hat p_s(a,x_s^c(a),j)-p_s(a;e_j)\right|
\le\beta_K\sqrt{x_s^c(a)^\top(V^c_{s,a})^{-1}x_s^c(a)}.
\label{eq:app-cost-confidence}
\end{gather}
Set $\alpha_c=\beta_K$, write $x=x_s^c(a)$, and abbreviate the clipped bound before monotone projection as $\hat U_s^c(a,j)$. On this event, \eqref{eq:cost-ucb} gives $0\le\hat U_s^c(a,j)-p_s(a;e_j)\le2\beta_K\sqrt{x^\top(V^c_{s,a})^{-1}x}$. For $i\le j$, $p_s(a;e_j)\le p_s(a;e_i)\le\hat U_s^c(a,i)$. The cumulative minimum therefore preserves coverage and can only reduce the excess at edge $j$.

If the smallest eigenvalue satisfies $\lambda_{\min}(V^c_{s,a})\ge\kappa n_{s,a}$ at every step considered, for fixed $\kappa>0$, a valid calibration in \eqref{eq:k-confidence-scale} is
\begin{equation}
\begin{aligned}
A&=1+\frac{d_c}{2}\log\!\left(1+\frac{NL_c^2}{\gamma_cd_c}\right)
+\log\!\left(\frac{M}{\delta_{\rm stat}}\right),\\
C_c&=\frac{2L_c}{\sqrt\kappa}
\left(\sqrt2R_c+\sqrt{\gamma_c}S_c\right).
\end{aligned}
\label{eq:app-calibration-constants}
\end{equation}
Substitution in \eqref{eq:exact-cost-radius} gives $\beta_K\le(\sqrt2R_c+\sqrt{\gamma_c}S_c)\sqrt{A+K\log2}$ since $A+K\log2>1$. Combining this with $\|x\|\le L_c$ and the eigenvalue bound gives
\[
0\le U_s^c(a,j)-p_s(a;e_j)
\le\frac{2\beta_K L_c}{\sqrt{\kappa n_{s,a}}}
\le C_c\sqrt{\frac{A+K\log2}{n_{s,a}}}=r_{s,K}.
\]
The factor $2$ covers prediction error and confidence bonus. For fixed $K$, the theorem therefore holds on this event at all pairs with $1\le n_{s,a}\le N$ satisfying its grid, margin, core, and design conditions. Here $\delta_{\rm stat}$ controls estimation, whereas $\delta_c$ bounds the next call's conditional tail probability. For finitely many candidate $K$ values, split failure probability equally across them and add their log count to $A$.

\subsection{Order Tightness}\label{app:k-tightness}
The following exact-core construction ($\eta=0$) attains both $K$-dependent orders in \eqref{eq:k-tradeoff-explicit} under the stated calibration.

\begin{proposition}[Order tightness]
\label{prop:k-tightness}
Fix $\delta_{\rm stat},\zeta\in(0,1)$ and integers $n,K\ge1$, with confidence parameter $\delta_{\rm stat}$ and finite-sample failure probability $\zeta$. The parameter $\delta_{\rm stat}$ enters the confidence radius used by the algorithm, whereas $\zeta$ controls the probability of the empirical-cdf event on which the finite-sample conclusion below holds. At horizon $N=n$, take independent normalized lengths $Y_i\sim\operatorname{Unif}[0,1]$ ($1\le i\le n$) from one model, constant feature $x=1$, unit ridge regularization, $\delta_c=1/2$, and
\[
\alpha_c=\beta_{n,K}
=\frac12\sqrt{\log(1+n)+2\log\!\left(\frac{2^K}{\delta_{\rm stat}}\right)}+1.
\]
Set $J=2^K$, $J_c=J-1$, $h=3/(4J_c)$, core edges $e_j=jh$ ($0\le j\le J_c$), and $e_J=1$. After $n$ observations, abbreviate the first-safe-edge length and cost by $\widehat q,\widehat c^+$. The target is $q=1/2$, with cost $c^{1/2}$ from \eqref{eq:true-safe-cost}. For price $\rho^{\rm out}>0$, $\widehat c^+-c^{1/2}=\rho^{\rm out}(\widehat q-1/2)$.

The calibration uses $M=d_c=L_c=S_c=\gamma_c=\kappa=1$, $R_c=1/2$, and $N=n$. Equation~\eqref{eq:app-calibration-constants} gives $A=A_n$, $C_c=C_0$, and the allowance and empirical-distribution radius
\[
\begin{aligned}
A_n&=1+\tfrac12\log(1+n)+\log(1/\delta_{\rm stat}),
& C_0&=2+\sqrt2,\\
r_{n,K}&=C_0\sqrt{\frac{A_n+K\log2}{n}},
& \varepsilon_n&=\sqrt{\frac{\log(2/\zeta)}{2n}}.
\end{aligned}
\]
If $\varepsilon_n+1/n\le\beta_{n,K}/(2\sqrt n)$ and $r_{n,K}\le1/4$, then with probability at least $1-\zeta$, Theorem~\ref{thm:ktradeoff}(i)--(iii) hold with $r_{s,K}=r_{n,K}$, $L_Q=3/4$, $\eta=0$, $\nu=1$, and
\begin{equation}
\frac{\rho^{\rm out}}6\left(h+\frac{\beta_{n,K}}{\sqrt n}\right)
\le \widehat c^+-c^{1/2}
\le 2\rho^{\rm out}\left(h+\frac{\beta_{n,K}}{\sqrt n}\right).
\label{eq:k-tightness}
\end{equation}
For fixed $\delta_{\rm stat}$ and $n\to\infty$, take $K=K(n)$ with $\log n=o(K)$ and $K=o(n)$. Then $\log n/K\to0$, $K/n\to0$, and $(\widehat c^+-c^{1/2})/[\rho^{\rm out}\sqrt{K/n}]\xrightarrow{P}\sqrt{\log2/2}$, where $\xrightarrow{P}$ denotes convergence in probability.
\end{proposition}

The comparison $\sqrt{(A_n+K\log2)/2}\le\beta_{n,K}\le(1+1/\sqrt2)\sqrt{A_n+K\log2}$ and $h=3/[4(2^K-1)]$, $\eta=0$, show that \eqref{eq:k-tightness} matches both $K,n$ orders in \eqref{eq:k-tradeoff-explicit}. This result retains $A_n$ because $N=n$ grows and applies only to the stated calibration.

\noindent\emph{Proof of Proposition~\ref{prop:k-tightness}.}
The scalar model has $p(e)=1-e$ on $[0,1]$, $\theta_j=1-e_j$ with $|\theta_j|\le1$, conditionally $1/2$-sub-Gaussian centered Bernoulli noise, and design $V_n=n+1\ge n$. The curve $\mathsf Q(u)=3u/4$ gives $L_Q=3/4$ and $\eta=0$. Since $p(q+u)=1/2-u$ for $0\le u\le1/2$, we also have $\nu=1$ and $u_0=1/2$.

Write $t=\beta_{n,K}/\sqrt n$, ridge bonus $b=\beta_{n,K}/\sqrt{n+1}$, shrinkage $\omega=n/(n+1)$, and $r=r_{n,K}$. Equation~\eqref{eq:app-calibration-constants} gives $2t\le r\le1/4$, so $t\le1/8$. Let $\widehat F_n$ be the empirical cdf, $\widehat S_n=1-\widehat F_n$, and $\mathcal E_n=\{\sup_{0\le u\le1}|\widehat F_n(u)-u|\le\varepsilon_n\}$.
The Dvoretzky--Kiefer--Wolfowitz inequality \citep{massart1990tight} gives $\Pr(\mathcal E_n)\ge1-\zeta$. Since $\widehat p_n(e)=\omega\widehat S_n(e)$ and $1-\omega=\omega/n$, on $\mathcal E_n$ the excess before clipping satisfies
\[
\begin{aligned}
\widehat p_n(e)+b-p(e)
&\ge b-\omega(\varepsilon_n+1/n)
\ge t(\sqrt\omega-\omega/2)>0,\\
\widehat p_n(e)+b-p(e)
&\le b+\omega\varepsilon_n\le3t/2.
\end{aligned}
\]
Clipping to $[0,1]$ preserves coverage and excess. The clipped bounds are nonincreasing in $e$, so the cumulative minimum leaves them unchanged. Hence $0\le U_n(e_j)-p(e_j)\le3t/2\le r$. Also $r\le1/4$ gives $r/\nu\le u_0$ and $q+r/\nu\le e_{J_c}=3/4$, verifying Theorem~\ref{thm:ktradeoff}(i)--(iii) on $\mathcal E_n$.

The safe-edge test requires $\widehat F_n(e)\ge1/2+\Delta$, where $\Delta=(n+1)b/n-1/(2n)=t\sqrt{1+1/n}-1/(2n)$. Thus on $\mathcal E_n$, the selected edge satisfies
\[
\widehat q-q\ge\Delta-\varepsilon_n
\ge t\bigl(\sqrt{1+1/n}-1/2\bigr)+\frac1{2n}
>\frac t2.
\]
Also, $q/h=2J_c/3$, so every grid edge strictly above $q$ is at least $h/3$ above it. Hence $\widehat q-q\ge\max\{h/3,t/2\}\ge(h+t)/6$.
For the upper bound, the first core edge at or above $v=q+3t/2<3/4$ lies within $h$ and has reported bound at most $p(v)+3t/2=1/2$. It qualifies, giving $\widehat q-q\le h+3t/2\le2(h+t)$. Cancelling input charge and multiplying by $\rho^{\rm out}$ proves \eqref{eq:k-tightness} on $\mathcal E_n$.

Set $\tau_n=1/2+\Delta=1/2+b+(b-1/2)/n$ and $m=\lceil n\tau_n\rceil=\lceil(n-1)/2+(n+1)b\rceil$. The test is $\widehat F_n(e)\ge\tau_n$. Since $b\le1/8$, we have $0<\tau_n<1$ and $1\le m\le n$. Thus, for $E_K=\{e_1,\ldots,e_J\}$ and the $m$th order statistic $Y_{(m)}$, $\widehat q=\min\{e\in E_K:e\ge Y_{(m)}\}$ almost surely. Also,
\[
\begin{aligned}
\mathbb E[Y_{(m)}]&=\frac{m}{n+1}=\frac12+b+O(n^{-1}),\\
\operatorname{Var}(Y_{(m)})
&=\frac{m(n+1-m)}{(n+1)^2(n+2)}\le\frac1{4(n+2)}.
\end{aligned}
\]
Hence $Y_{(m)}=1/2+b+O_P(n^{-1/2})$. Under fixed $\delta_{\rm stat}$, $\log n=o(K)$, and $K=o(n)$, we have $\beta_{n,K}\sim\sqrt{K\log2/2}$, $b\sim\sqrt{K\log2/(2n)}$, and $h=o(\sqrt{K/n})$. With probability tending to one, $Y_{(m)}\le3/4$, so its next edge is in the core and $0\le\widehat q-Y_{(m)}<h$ almost surely. Thus $\widehat q-1/2=b+O_P(n^{-1/2})+O_P(h)$. Divide by $\sqrt{K/n}$ and use $\widehat c^+-c^{1/2}=\rho^{\rm out}(\widehat q-1/2)$ for the claimed limit. For fixed $\zeta\in(0,1)$, the finite-sample conditions eventually hold since $r\to0$ and $(\varepsilon_n+1/n)/t\to0$.\hfill$\square$

\subsection{Grid Conditions and Extension}\label{app:k-grid-extension}
A density bounded below by $1/L_Q$ throughout the reference core support gives $L_Q$-Lipschitz quantiles. To compare grids across $K$, fix the state, sample size, and $L_Q,C_c,A,\nu,u_0$, and require each grid to satisfy the theorem. The placement error $\eta$ may vary, but the two-term rate requires common $C_h$. Separate routing runs may use different data.

For fixed $(s,a)$, write $q=q_s(a)$ and $r=r_{s,K}$. A frozen, strictly increasing grid has upward rounding distance $\Delta_K(u)=\min_{j\in[J],\,e_j\ge u}(e_j-u)$ on $[0,e_J]$. The same proof under \eqref{eq:k-confidence-excess} and \eqref{eq:tail-margin} gives, if $r/\nu\le u_0$ and $q+r/\nu\le e_J$,
\[
0\le\hat c_s^+(a)-c_s^{\delta_c}(a)
\le\rho_a^{\rm out}\left[\Delta_K(q+r/\nu)+r/\nu\right].
\]
Equal-width grids give $\Delta_K(u)\le e_J2^{-K}$. Quantile meshes can have large tail rounding errors, so their core-resolution rate requires $q+r/\nu$ in the core.

The experiments reuse warm data for placement and fitting, tune $\alpha_c$, and use an empirical top edge. Because these choices need not satisfy the calibration assumptions, we measure coverage directly in \prettyref{app:calibration}.

\section{Implementation and Evaluation Details}
\label{app:repro}
\subsection{Learners, Frozen Mesh, and Parameters}
\label{app:params}

We use normalized BGE-small-en-v1.5 embeddings and seed-specific QR for projection $P$ in Section~\ref{sec:reward-module}. The cost feature $x_s^c(a)=[z_s;\xi_s(a)]$ has dimension $d_c=d_0+10$. Its ten pre-call components are retry fraction, retry-depth fraction, current-task-spend ratio, remaining-budget ratio, log input-token scale, candidate-arm price scale, previous-arm price scale, previous reward, previous-cost scale, and a same-as-previous-arm indicator. 

\paragraph{Ridge learners.}
Fitting includes warm calls ($u\le0$, $G_u=1$) and online calls before $s$. The reward feature $\Phi_s(a)=[z_s;0;\ldots;x_s^r;\ldots;0]\in\R^{D_r}$, $D_r=d_0+Md_r$, places $x_s^r$ in model $a$'s block. With ridge penalties $\gamma_r,\gamma_c>0$ and identity matrix $I_d$,
\[
V_s^r=\gamma_r I_{D_r}+\sum_{u<s:G_u=1}\Phi_u(a_u)\Phi_u(a_u)^\top,
\quad b_s^r=\sum_{u<s:G_u=1}r_u\Phi_u(a_u),
\quad\hat w_s=(V_s^r)^{-1}b_s^r.
\]
In \eqref{eq:hybrid-reward}, $\hat\beta_s$ and $\hat\theta_{s,a}$ are the first $d_0$ entries and subsequent model blocks of $\hat w_s$. We use variance proxy $v_s^r(a)=\Phi_s(a)^\top(V_s^r)^{-1}\Phi_s(a)$ and Schur complements with $\gamma_r=1$.
Cost heads share a design matrix with threshold-specific feature-label vectors,
\begin{equation}
\begin{aligned}
V^c_{s,a}&=\gamma_c I_{d_c}+\sum_{u<s:G_u=1,a_u=a}x_u^c(a)x_u^c(a)^\top,\\
b^c_{s,a,j}&=\sum_{u<s:G_u=1,a_u=a}\1\{\ell_u^{\rm out}>e_j\}x_u^c(a).
\end{aligned}
\label{eq:cost-design}
\end{equation}
Writing $x=x_s^c(a)$, their estimates and common variance proxy are
\[
\hat p_s(a,j)=x^\top(V^c_{s,a})^{-1}b^c_{s,a,j},\qquad
v_s^c(a)=x^\top(V^c_{s,a})^{-1}x.
\]

\paragraph{Warm mesh.}
Experiment~1 uses unconstrained Greedy LinUCB warm logs, fixes $\lambda,\phi$ during replay, and rebuilds cost heads on the finalized grid. Let the multiset $\mathcal L_w$ contain warm completion-token counts (estimated from cost if unavailable) and $Q_\tau(A)$ be the empirical $\tau$-quantile. With one tail and $J_c=2^K-1$ core intervals, set $e_{\rm core}=\max\{1,Q_{\tau_{\rm cover}}(\mathcal L_w)\}$,
where $\tau_{\rm cover}=0.95$ for \textsc{MATH500} and $0.75$ for \textsc{GPQA}. For $\ell^{\rm clip}=\min\{\max\{\ell,0\},e_{\rm core}\}$, place $e_m=Q_{m/J_c}(\{\ell^{\rm clip}:\ell\in\mathcal L_w\})$ for $m=0,\ldots,J_c$, with $e_0=0$ and $e_{J_c}=e_{\rm core}$. Replace each nonincreasing edge by its predecessor plus $10^{-6}$, which can move $e_{J_c}$ slightly above the target $e_{\rm core}$. The top edge is
\[
e_J=\max\{4e_{\rm core},\;1.05\max\mathcal L_w,\;e_{\rm core}+1\}.
\]
Below $\max\{32,J_c+1\}$ warm samples, we use a geometric core grid. All reported runs permit quantile placement. Length labels use $K$ bits, while the ledger, controller, and features retain exact charges and prior outcomes.

\paragraph{Execution details.}
Score ties favor reward UCB, then the smaller mean-cost proxy. For fixed $s,a$, let $\bar p_j=\min_{1\le i\le j}\clip_{[0,1]}\hat p_s(a,i)$ and $\bar p_0=1$.
The left-edge approximation to the survival integral on $[0,e_J]$ gives
\[
\hat\mu_s^c(a)=\rho_a^{\rm in}\ell_s^{\rm in}(a)+\rho_a^{\rm out}\sum_{j=1}^J(e_j-e_{j-1})\bar p_{j-1}.
\]
We follow Algorithm~\ref{alg:adpht_brief} with pacing after four samples and stopping rule \eqref{eq:tau}. Spending discrepancies in \eqref{eq:lambda-eg} and \eqref{eq:phi-eg} are clipped to $[-1,4]$, and exponents to $[-10,10]$. Neither limit binds. Table~\ref{tab:app-adp-profile} gives dataset-specific settings, with ``init'' denoting initial online values. Budget-aware (BA) and Position-aware (PA) LinUCB share parameters.

\begin{table}[!htbp]
\centering
\caption{Experiment~1 implementation settings.}
\label{tab:app-adp-profile}
\scriptsize
\begin{tabular}{lcc}
\toprule
Setting & \textsc{MATH500} & \textsc{GPQA} \\
\midrule
Seeds, warm/test split & $42$--$46$, $20\%/80\%$ & $42$--$46$, $20\%/80\%$ \\
Warm-start, retry upper bound & Greedy infinite-budget replay, $H=4$ & Greedy infinite-budget replay, $H=4$ \\
Embedding and grader & BGE-small-en-v1.5, GPT-OSS-120B & BGE-small-en-v1.5, GPT-OSS-120B \\
Arms & \multicolumn{2}{c}{Qwen3-8B, Phi-4, Gemini-Flash-Lite, DeepSeek-V3.2, GPT-5-mini} \\
Budgets (USD) & $0.025,0.05,0.075,0.125,0.30,0.45,0.60$ & MATH grid plus $0.80,1.00$ \\
\midrule
$K$, mesh edges, $d_0$ & $2$, warm-start quantiles, $8$ & $2$, warm-start quantiles, $8$ \\
Core coverage quantile & $0.95$ & $0.75$ \\
$\alpha_r$, $\alpha_c$, $\delta_c$ & $0.25$, $0.35$, $0.10$ & $0.35$, $0.18$, $0.24$ \\
$\lambda_{\rm init}$, $\lambda_{\max}$, $\eta_\lambda$ & $0.25$, $8.0$, $0.05$ & $0.50$, $12.0$, $0.08$ \\
$\lambda_{\min}$ & $10^{-4}$ & $10^{-4}$ \\
$\phi_{\rm init}$, $[\phi_{\min},\phi_{\max}]$, $\eta_\phi$ & $1.0$, $[0.5,3.0]$, $0.12$ & $0.80$, $[0.35,2.5]$, $0.10$ \\
$\kappa_0$, $\kappa_h$, $\kappa_b$ & $0.02$, $0.10$, $0.50$ & $0.03$, $0.12$, $0.65$ \\
\bottomrule
\end{tabular}
\end{table}

\subsection{Experiment 1 Protocol and Baselines}
\label{app:eval-protocol}

Stratified 20\%/80\% splits give 100/400 warm/test problems for \textsc{MATH500} and 90/358 for \textsc{GPQA}. Policies share warm logs, task order, prompts, GPT-OSS-120B verifier, local BGE embeddings \citep{baai2026bge}, model menu, retry limit, and ledger. Tables~\ref{tab:app-adp-profile} and~\ref{tab:app-model-prices} give seeds, budgets, and token prices.

Calls use one user message and no system prompt \citep{poon2026multillm}. The first-call template is
\begin{lstlisting}
Solve the following problem: {question}
Please answer in a short, clear, and concise way. Show only the essential derivation, and end with the final answer in LaTeX format, enclosed within \boxed{}. For example: 'The final answer is \boxed{x=5}'.
\end{lstlisting}
After failure, we retain the question and append failed attempts in order.
\begin{lstlisting}
The previous attempts below were incorrect. Please answer the problem again with the same short, clear, and concise style.
-----------------------------------------------------
- Attempt 1 (by {model_id_1}): {failed_response_1}
- Attempt 2 (by {model_id_2}): {failed_response_2}
...
-----------------------------------------------------
Now please provide your new answer based on the above attempts and the original question.
\end{lstlisting}
The grader sees only candidate and reference answers. Reward is one exactly when its whitespace-stripped output is \texttt{True}.
\begin{lstlisting}
Expression 1: {model_answer}
Expression 2: {ground_truth_answer}

Expression 2 is the answer and expression 1 is attempt by student. Look at their final answer only, which might be boxed. Does the student get the final expected answer? Respond with only the word 'True' or 'False'.
\end{lstlisting}

\paragraph{Implemented baselines.}
LinUCB baselines fit rewards by arm. Greedy selects by reward UCB, and its unconstrained trajectory is replayed under \eqref{eq:tau}. BA admits arms whose empirical mean cost plus one standard deviation fits $R_s/N_s$, then ranks reward UCB divided by a cost lower confidence bound. PA solves a 0--1 knapsack with these cost estimates and reward UCBs, choosing the highest reward-UCB feasible arm. Both use exploration scale $0.675$, cost coefficient $1.0$, and ridge penalty $0.45$ on both datasets \citep{poon2026multillm}.

ParetoBandit \citep{tabernermiller2026paretobandit} uses exploration scale $\alpha=1.0$, variance-inflation cap $v_{\max}=200$, geometric forgetting $\gamma=0.997$, static cost weight $\lambda_c=0.3$, adaptive dual cap $\bar\lambda=5.0$, dual step $\eta=0.05$, EMA coefficient $0.05$, cost floor $10^{-7}$, no forced initial pulls, and finite-stream average-cost target $R_s/(N_sH)$. Preference and cascade routers lack this replay protocol and appear as related systems.

\begin{table}[!htbp]
\centering
\caption{Model identifiers and configured token prices (USD/token).}
\label{tab:app-model-prices}
\footnotesize
\begin{tabular}{lcc}
\toprule
Model & Input price & Output price \\
\midrule
qwen/qwen3-8b & $5.0\times10^{-8}$ & $4.0\times10^{-7}$ \\
microsoft/phi-4 & $6.5\times10^{-8}$ & $1.4\times10^{-7}$ \\
google/gemini-2.5-flash-lite & $1.0\times10^{-7}$ & $4.0\times10^{-7}$ \\
deepseek/deepseek-v3.2 & $2.6\times10^{-7}$ & $3.8\times10^{-7}$ \\
openai/gpt-5-mini & $2.5\times10^{-7}$ & $2.0\times10^{-6}$ \\
\bottomrule
\end{tabular}
\end{table}

\subsection{RouterBench Protocol and Configurations}
\label{app:routerbench-protocol}

\textsc{RouterBench} has 36,497 prompts and eleven models, with scores in $[0,1]$ (78.9\% binary). Seeds 42--46 give task-stratified 20\%/80\% warm/test splits of approximately 7,298/29,199 prompts. We set $H=1$, reveal each selected outcome, and charge its stored cost under the common crossing rule. Cumulative score sums outcomes until stopping. Completion length is estimated as $\max\{1,\text{characters}/4\}$ from stored responses. Input token price is zero. Output token price is each model's median cost divided by its median proxy length. 

All five policies share Greedy chosen-arm warm logs, splits, test order, and ledger across budgets, with baselines from \prettyref{app:eval-protocol}.  Table~\ref{tab:app-routerbench-native} reports five-seed means to one decimal.

\begin{table}[!htbp]
\centering
\caption{RouterBench cumulative scores, with five-seed means and budgets in USD.}
\label{tab:app-routerbench-native}
\footnotesize
\setlength{\tabcolsep}{4.5pt}
\begin{tabular}{rccccc}
\toprule
Budget & \algname{} & BA & PA & ParetoBandit & Greedy \\
\midrule
$0.500$ & $2{,}745.2$ & $2{,}083.4$ & $2{,}192.2$ & $1{,}714.7$ & $330.8$ \\
$1.000$ & $5{,}616.6$ & $4{,}120.2$ & $4{,}720.7$ & $2{,}564.3$ & $1{,}137.2$ \\
$1.500$ & $9{,}827.2$ & $6{,}601.7$ & $6{,}982.4$ & $3{,}389.5$ & $1{,}283.0$ \\
$2.000$ & $11{,}142.8$ & $8{,}234.4$ & $8{,}664.6$ & $4{,}242.6$ & $1{,}361.9$ \\
$2.500$ & $12{,}157.7$ & $9{,}309.8$ & $9{,}891.7$ & $5{,}992.2$ & $1{,}545.0$ \\
$3.500$ & $13{,}731.9$ & $13{,}375.3$ & $13{,}586.9$ & $9{,}041.4$ & $1{,}857.7$ \\
$4.400$ & $14{,}737.0$ & $13{,}013.4$ & $14{,}423.9$ & $12{,}794.8$ & $1{,}988.4$ \\
$5.336$ & $15{,}323.9$ & $13{,}330.8$ & $14{,}831.2$ & $14{,}292.5$ & $2{,}116.2$ \\
$7.000$ & $16{,}815.8$ & $13{,}677.8$ & $17{,}293.8$ & $14{,}727.4$ & $2{,}347.0$ \\
$8.858$ & $17{,}702.6$ & $14{,}135.3$ & $18{,}122.0$ & $15{,}030.9$ & $2{,}608.7$ \\
$13.340$ & $18{,}496.9$ & $14{,}814.1$ & $18{,}295.6$ & $15{,}390.6$ & $3{,}197.4$ \\
$26.681$ & $18{,}718.4$ & $14{,}667.8$ & $18{,}813.8$ & $15{,}771.6$ & $4{,}892.3$ \\
$53.361$ & $20{,}101.2$ & $15{,}100.8$ & $19{,}229.2$ & $15{,}847.6$ & $19{,}534.0$ \\
$80.042$ & $20{,}200.2$ & $15{,}020.0$ & $19{,}463.6$ & $15{,}893.2$ & $20{,}340.7$ \\
\bottomrule
\end{tabular}

\end{table}

\subsection{Commercial-Router Protocol}
\label{app:commercial-protocol}

The seed-42 split of 1,055 \textsc{LiveCodeBench} release\_v6 problems gives 211 warm and 844 test problems. An isolated grader requires all public and private tests to pass. Requests use temperature zero, one call per problem, and no output-token cap. All routers use the same 39 candidates from July~1, 2026 without output-cap filtering.

Default and cost-saving AutoRouter use \texttt{cost\_quality\_tradeoff} 7 and 10, respectively. Each full pass is reordered to the common sequence and replayed under \eqref{eq:tau} at every budget. Each \algname{} budget has a separate live run.

Table~\ref{tab:app-commercial} counts \algname{} errors as unsolved with zero reported charge. Default AutoRouter solves 623 for $\$22.86$ (one error, 32 truncation flags), and cost-saving solves 732 for $\$1.62$ (seven errors, 14 flags). Auto-CS and Auto-default denote these settings, with default spending in parentheses. Frequent errors complicate comparison at larger budgets.

\begin{table}[!htbp]
\centering
\caption{Commercial comparison on the common 844-problem \textsc{LiveCodeBench} order.}
\label{tab:app-commercial}
\footnotesize
\setlength{\tabcolsep}{3pt}
\begin{tabular}{rrrrrrr}
\toprule
Nominal $B$ & \algname{} solved & \algname{} spend & Errors & Auto-CS solved & Auto-CS spend & Auto-default (spend) \\
\midrule
$0.202$ & 244 & $0.2024$ & 0 & 86 & $0.2020$ & 8 ($0.5013$) \\
$0.404$ & 436 & $0.4041$ & 0 & 166 & $0.4116$ & 8 ($0.5013$) \\
$0.8081$ & 682 & $0.8084$ & 0 & 372 & $0.8092$ & 15 ($0.8190$) \\
$5.710$ & 704 & $5.7103$ & 1 & 732 & $1.6161$ & 154 ($5.7851$) \\
$11.430$ & 628 & $7.4154$ & 137 & 732 & $1.6161$ & 298 ($11.4701$) \\
$17.140$ & 710 & $15.5118$ & 36 & 732 & $1.6161$ & 420 ($17.3688$) \\
$22.860$ & 615 & $10.7744$ & 137 & 732 & $1.6161$ & 623 ($22.8551$) \\
\bottomrule
\end{tabular}
\end{table}

\subsection{Experiment 1 Results and Uncertainty}
\label{app:tables}

Table~\ref{tab:app-main} gives five-seed means and sample standard deviations for Figure~\ref{fig:frontier}. B-aware, P-aware, and Pareto denote BA, PA, and ParetoBandit. Gap is \algname{}'s mean minus the best of these baseline means. The plotted 95\% intervals use 10,000 bootstrap resamples with a deterministic hash seed. Budgets share warm logs, making outcomes dependent.

Table~\ref{tab:app-hierboot} reports mean differences in solved counts, with equal weight on each budget in the fixed grid. For each dataset, we construct approximate paired 95\% two-level bootstrap confidence intervals. Each draw first resamples budget cells with replacement and then resamples the five seed-level policy pairs within each drawn budget, keeping \algname{} and the comparator aligned on the same seed. The statistic is the equally weighted mean difference over the fixed budget grid. The comparisons also reflect implementation and tuning differences.

Table~\ref{tab:operational-paths} describes spending. Utilization averages spending/budget on \textsc{MATH500}. Profiles pool five \algname{} runs at the largest budget, one per task/replicate. Unsolved is their unsolved fraction, and Unsolved spend is those tasks' spending share. Mean costs separate solved and unsolved tasks. Unsolved tasks cost more in both datasets, motivating control of retry spending.

\begin{table}[!htbp]
\centering
\caption{Experiment~1 solved counts (mean $\pm$ sample standard deviation over five seeds). Budgets are in USD.}
\label{tab:app-main}
\setlength{\tabcolsep}{1.6pt}
\footnotesize
\begin{tabular}{llcccccc}
\toprule
Dataset & Budget & Greedy & B-aware & P-aware & Pareto & \algname{} & Gap \\
\midrule
\textsc{MATH500} & 0.025 & 31.0 $\pm$ 14.8 & 96.6 $\pm$ 8.0 & 100.8 $\pm$ 16.8 & 79.6 $\pm$ 13.0 & 216.2 $\pm$ 83.8 & 115.4 \\
 & 0.050 & 53.0 $\pm$ 18.8 & 197.0 $\pm$ 22.0 & 191.0 $\pm$ 21.6 & 151.0 $\pm$ 23.3 & 261.4 $\pm$ 67.7 & 64.4 \\
 & 0.075 & 69.4 $\pm$ 19.8 & 263.4 $\pm$ 18.8 & 257.8 $\pm$ 23.1 & 252.2 $\pm$ 33.6 & 300.4 $\pm$ 29.1 & 37.0 \\
 & 0.125 & 103.0 $\pm$ 30.9 & 313.8 $\pm$ 7.7 & 329.2 $\pm$ 8.0 & 329.6 $\pm$ 30.9 & 339.0 $\pm$ 22.4 & 9.4 \\
 & 0.300 & 249.8 $\pm$ 67.5 & 332.4 $\pm$ 13.5 & 369.4 $\pm$ 5.8 & 370.8 $\pm$ 8.8 & 371.4 $\pm$ 17.5 & 0.6 \\
 & 0.450 & 353.8 $\pm$ 76.6 & 344.6 $\pm$ 23.0 & 381.2 $\pm$ 3.4 & 380.6 $\pm$ 7.0 & 386.2 $\pm$ 5.3 & 5.0 \\
 & 0.600 & 375.4 $\pm$ 33.3 & 355.8 $\pm$ 19.0 & 384.4 $\pm$ 5.0 & 384.4 $\pm$ 6.2 & 387.0 $\pm$ 2.1 & 2.6 \\
\midrule
\textsc{GPQA} & 0.025 & 10.4 $\pm$ 1.5 & 28.0 $\pm$ 4.8 & 29.8 $\pm$ 4.7 & 26.8 $\pm$ 3.6 & 85.0 $\pm$ 4.8 & 55.2 \\
 & 0.050 & 16.2 $\pm$ 2.8 & 54.4 $\pm$ 10.4 & 57.0 $\pm$ 9.7 & 57.6 $\pm$ 8.1 & 132.4 $\pm$ 8.4 & 74.8 \\
 & 0.075 & 23.0 $\pm$ 4.8 & 90.2 $\pm$ 8.0 & 90.2 $\pm$ 9.0 & 93.2 $\pm$ 4.9 & 147.4 $\pm$ 9.8 & 54.2 \\
 & 0.125 & 38.8 $\pm$ 4.7 & 142.2 $\pm$ 10.6 & 137.4 $\pm$ 7.2 & 150.8 $\pm$ 6.1 & 184.2 $\pm$ 10.8 & 33.4 \\
 & 0.300 & 90.6 $\pm$ 11.6 & 203.0 $\pm$ 11.2 & 180.0 $\pm$ 19.7 & 235.2 $\pm$ 21.1 & 242.0 $\pm$ 7.4 & 6.8 \\
 & 0.450 & 133.0 $\pm$ 14.7 & 229.6 $\pm$ 33.7 & 230.2 $\pm$ 16.1 & 249.0 $\pm$ 7.4 & 254.4 $\pm$ 6.4 & 5.4 \\
 & 0.600 & 174.6 $\pm$ 20.8 & 246.2 $\pm$ 22.9 & 262.4 $\pm$ 7.8 & 260.0 $\pm$ 5.0 & 265.6 $\pm$ 8.8 & 3.2 \\
 & 0.800 & 240.4 $\pm$ 31.4 & 239.6 $\pm$ 28.9 & 263.8 $\pm$ 8.9 & 258.4 $\pm$ 2.1 & 266.4 $\pm$ 10.8 & 2.6 \\
 & 1.000 & 269.4 $\pm$ 10.6 & 250.4 $\pm$ 14.3 & 258.6 $\pm$ 2.5 & 264.0 $\pm$ 8.4 & 268.6 $\pm$ 10.6 & 4.6 \\
\bottomrule
\end{tabular}
\end{table}

\begin{table}[!htbp]
\centering
\caption{Experiment~1 hierarchical-bootstrap gaps over each dataset's budget grid.}
\label{tab:app-hierboot}
\footnotesize
\begin{tabular}{llccc}
\toprule
Dataset & Baseline & Mean diff. & $95\%$ CI & $p$-value (one-sided) \\
\midrule
\textsc{GPQA} & Budget-aware & $40.3$ & [27.4,\;54.2] & $<10^{-4}$ \\
\textsc{GPQA} & Position-aware & $37.4$ & [19.6,\;54.2] & $<10^{-4}$ \\
\textsc{GPQA} & ParetoBandit & $27.9$ & [11.3,\;46.3] & $<10^{-4}$ \\
\textsc{MATH500} & Budget-aware & $51.1$ & [31.5,\;80.4] & $<10^{-4}$ \\
\textsc{MATH500} & Position-aware & $35.4$ & [7.1,\;71.3] & $0.001$ \\
\textsc{MATH500} & ParetoBandit & $44.8$ & [8.1,\;88.2] & $0.002$ \\
\bottomrule
\end{tabular}

\end{table}

\begin{table}[!htbp]
\centering
\caption{Experiment~1 operational-path summaries from the five-seed main-frontier runs.}
\label{tab:operational-paths}
\footnotesize
\setlength{\tabcolsep}{3.5pt}
\begin{tabular}{lrrrr}
\toprule
\multicolumn{5}{l}{Panel A, mean budget utilization on \textsc{MATH500} (\%)} \\
Policy & \algname{} & BA & PA & ParetoBandit \\
\midrule
Realized spend / nominal budget & 93.4 & 75.7 & 77.2 & 81.5 \\
\midrule
\multicolumn{5}{l}{Panel B, maximum-budget task profiles for \algname{}} \\
Dataset ($B$) & Profiles & Unsolved (\%) & Unsolved spend (\%) & Mean cost, solved / unsolved \\
\midrule
\textsc{MATH500} ($0.60$) & 2,000 & 3.25 & 20.2 & $0.926 / 6.989$ \\
\textsc{GPQA} ($1.00$) & 1,784 & 24.72 & 53.9 & $1.331 / 4.732$ \\
\bottomrule
\end{tabular}
\par\vspace{1mm}
\scriptsize Mean task costs in Panel~B are in $10^{-3}$ USD.
\end{table}

\subsection{Resolution Sweep}
\label{app:kablation}

We vary $K\in\{1,\ldots,5\}$ on both datasets at $B=0.05,0.30$, with five replicates per setting. Each replicate shares warm lengths and the protocol of \prettyref{app:eval-protocol} across $K$. These runs are separate from the main frontier. Default $K=2$ predates this analysis.

Table~\ref{tab:kablation} reports solved counts (Panel~A) and paired two-level bootstrap with equal-weight settings (Panel~B). Intervals favor $K=2$ over $K=4,5$ but include zero against $K=1,3$. Provider variation and routing history also affect outcomes, so this does not establish an optimal $K$ or verify the theorem's assumptions.

\begin{table}[!htbp]
\centering
\caption{Resolution sweep with five-seed solved counts (mean $\pm$ sample standard deviation) and pooled two-level bootstrap comparisons. Budgets are in USD.}
\label{tab:kablation}
\label{tab:kboot}
\footnotesize
\begingroup
\setlength{\tabcolsep}{4pt}
\textit{Panel A. Cell-level solved counts}\par\smallskip
\begin{tabular}{llccccc}
\toprule
Dataset & Budget & $K=1$ & $K=2$ & $K=3$ & $K=4$ & $K=5$ \\
\midrule
\textsc{MATH500} & 0.05 & 273.8 $\pm$ 49.3 & 270.2 $\pm$ 67.8 & 265.4 $\pm$ 75.3 & 266.4 $\pm$ 75.7 & 245.8 $\pm$ 83.8 \\
\textsc{MATH500} & 0.3 & 370.4 $\pm$ 17.9 & 371.0 $\pm$ 16.2 & 367.0 $\pm$ 19.4 & 334.4 $\pm$ 88.3 & 335.0 $\pm$ 91.0 \\
\textsc{GPQA} & 0.05 & 124.6 $\pm$ 11.7 & 138.6 $\pm$ 13.2 & 135.0 $\pm$ 6.7 & 131.8 $\pm$ 10.5 & 129.4 $\pm$ 9.8 \\
\textsc{GPQA} & 0.3 & 232.8 $\pm$ 15.2 & 243.2 $\pm$ 3.9 & 241.8 $\pm$ 9.8 & 239.8 $\pm$ 7.3 & 240.0 $\pm$ 7.2 \\
\bottomrule
\end{tabular}

\endgroup
\par\smallskip
\textit{Panel B. Two-level bootstrap over the four cells}\par\smallskip
\begin{tabular}{lcc}
\toprule
Comparison & Mean diff. & $95\%$ CI \\
\midrule
$K=2$ minus $K=1$ & $5.3$ & [$-4.5,\;14.1$] \\
$K=2$ minus $K=3$ & $3.5$ & [$-1.0,\;8.2$] \\
$K=2$ minus $K=4$ & $12.7$ & [$0.7,\;38.9$] \\
$K=2$ minus $K=5$ & $18.2$ & [$1.2,\;46.3$] \\
\bottomrule
\end{tabular}

\end{table}

\subsection{Coverage of the Deployed Cost Bound}
\label{app:calibration}

We check coverage empirically because the deployed grid and heads share warm data and use tuned confidence scales. Table~\ref{tab:calibration} compares selected-arm bounds $\hat c_s^+$ with provider-reported charges $c_s$ over two budgets and five seeds per dataset and $K$ (ten runs), excluding gate skips. Coverage is the fraction $c_s\le\hat c_s^+$. All-call gap averages $\max\{c_s-\hat c_s^+,0\}$ over all calls ($10^{-5}$ USD). Conditional severity averages $c_s-\hat c_s^+$ over exceedances ($10^{-4}$ USD). Exceedance above the target core cutoff is the fraction of raw completion-token counts above $e_{\rm core}$.

At $K=2$, coverage is 0.876 on \textsc{GPQA} and 0.988 on \textsc{MATH500}, above their respective targets $1-\delta_c=0.76,0.90$. Coverage declines as $K$ increases. On \textsc{GPQA}, it falls below target at $K=4,5$, reaching 0.751 and 0.715, while \textsc{MATH500} remains above target at $K=5$. For \textsc{MATH500} at $K=1$, coverage rounds from 0.9998 to 1.000, which explains why a positive gap is still reported. These pooled summaries do not establish simultaneous conditional coverage or identify why coverage declines.

Core overflow uses the target $e_{\rm core}$, which duplicate-edge correction can place below the actual tail interval. It does not use top edge $e_J$.

\begin{table}[!htbp]
\centering
\caption{Executed-call cost-bound audit across $K$.}
\label{tab:calibration}
\footnotesize
\begin{tabular}{llccccc}
\toprule
Dataset & $K$ & cells & Coverage & All-call gap & Cond. severity & Target-core exceedance \\
 & & & & ($\times 10^{-5}$) & ($\times 10^{-4}$) & rate \\
\midrule
\textsc{GPQA} & 1 & 10 & 0.960 & 2.40 & 6.08 & 0.0126 \\
\textsc{GPQA} & 2 & 10 & 0.876 & 4.11 & 3.31 & 0.0137 \\
\textsc{GPQA} & 3 & 10 & 0.807 & 5.56 & 2.88 & 0.0124 \\
\textsc{GPQA} & 4 & 10 & 0.751 & 5.68 & 2.28 & 0.0116 \\
\textsc{GPQA} & 5 & 10 & 0.715 & 6.53 & 2.29 & 0.0112 \\
\textsc{MATH500} & 1 & 10 & 1.000 & 0.03 & 12.28 & 0.0006 \\
\textsc{MATH500} & 2 & 10 & 0.988 & 0.13 & 1.04 & 0.0015 \\
\textsc{MATH500} & 3 & 10 & 0.970 & 0.76 & 2.54 & 0.0010 \\
\textsc{MATH500} & 4 & 10 & 0.959 & 1.26 & 3.08 & 0.0009 \\
\textsc{MATH500} & 5 & 10 & 0.944 & 1.49 & 2.68 & 0.0005 \\
\bottomrule
\end{tabular}

\end{table}

\makeatletter
\long\def\@makecaption#1#2{%
  \vskip\abovecaptionskip
  \sbox\@tempboxa{#1. #2}%
  \ifdim\wd\@tempboxa>\hsize
    #1. #2\par
  \else
    \global\@minipagefalse
    \hb@xt@\hsize{\hfil\box\@tempboxa\hfil}%
  \fi
  \vskip\belowcaptionskip}
\makeatother

\section{Cost-Head Ablation on RouterBench}
\label{app:cost-head-exceedance}

We compare three cost heads within one routing controller. \textbf{K-Bit} is the $K$-bit threshold-survival head implemented in \algname{}. \textbf{Continuous Mean} replaces that head with a continuous output-length mean predictor. \textbf{Mean-Adjusted} adds a one-sided confidence correction to Continuous Mean.

\paragraph{Protocol.}
We use seeds $42$--$46$ and budgets $5.00$, $8.86$, $10.00$, $13.34$, $20.00$, $40.02$, and $80.04$ USD. Each seed applies the original$20\%/80\%$ split. All heads share the warm observations, reward learner, dual update, pacing, model order, and test order.

\begin{table}[!htbp]
\centering
\caption{K-Bit, Continuous Mean, and Mean-Adjusted on \textsc{RouterBench} (five seeds).}
\label{tab:cost-head-exceedance}
\footnotesize
\setlength{\tabcolsep}{4pt}
\begin{tabular}{lrrr}
\toprule
Budget (USD) & K-Bit & Continuous Mean & Mean-Adjusted \\
\midrule
\multicolumn{4}{l}{\emph{Panel A. Cumulative score, mean (sample standard deviation)}} \\
5.00 & \textbf{15,989.4 (770.9)} & 14,774.2 (768.9) & 15,502.3 (416.7) \\
8.86 & \textbf{18,397.6 (462.3)} & 16,923.7 (1,589.2) & 17,419.5 (1,048.8) \\
10.00 & \textbf{18,673.3 (376.0)} & 17,248.3 (1,359.2) & 17,446.2 (1,199.9) \\
13.34 & \textbf{18,931.1 (515.2)} & 17,851.0 (1,330.5) & 17,762.9 (1,310.1) \\
20.00 & \textbf{18,996.2 (532.9)} & 18,771.0 (883.9) & 18,668.4 (992.8) \\
40.02 & 19,888.3 (315.2) & \textbf{20,072.6 (151.7)} & 20,054.0 (188.7) \\
80.04 & 20,272.6 (187.2) & 20,263.1 (195.8) & \textbf{20,283.6 (184.1)} \\
\midrule
\multicolumn{4}{l}{\emph{Panel B. Mean number of queries served}} \\
5.00 & 29,199.0 & 27,564.4 & 28,720.4 \\
8.86 & 29,198.2 & 28,450.2 & 29,164.4 \\
10.00 & 29,198.8 & 28,704.4 & 29,078.4 \\
13.34 & 29,199.0 & 29,068.0 & 29,197.8 \\
20.00 & 29,199.0 & 29,199.0 & 29,199.0 \\
40.02 & 29,199.0 & 29,199.0 & 29,199.0 \\
80.04 & 29,199.0 & 29,199.0 & 29,199.0 \\
\midrule
\multicolumn{4}{l}{\emph{Panel C. Mean score per served query}} \\
5.00 & 0.5476 & 0.5361 & 0.5397 \\
8.86 & 0.6301 & 0.5940 & 0.5972 \\
10.00 & 0.6395 & 0.6004 & 0.5998 \\
13.34 & 0.6483 & 0.6139 & 0.6084 \\
20.00 & 0.6506 & 0.6429 & 0.6394 \\
40.02 & 0.6811 & 0.6874 & 0.6868 \\
80.04 & 0.6943 & 0.6940 & 0.6947 \\
\bottomrule
\end{tabular}
\par\smallskip
\begin{minipage}{0.98\linewidth}\footnotesize
\end{minipage}
\end{table}

Table~\ref{tab:cost-head-exceedance} shows that K-Bit leads at the tight and intermediate budgets, while Continuous Mean and Mean-Adjusted become more competitive as the budget expands. At the tightest budget, K-Bit reaches the end of the test stream, whereas the continuous heads exhaust their budgets earlier. It also achieves higher average quality per served query, so the gain reflects both broader service and the models selected. Once all three heads reach the full stream, allocation quality rather than service reach determines the comparison. K-Bit retains the strongest per-query performance through the intermediate budgets. Continuous Mean has the strongest per-query performance at the next-to-highest budget, and Mean-Adjusted holds a slight advantage at the highest budget.

\section{Validation and Analysis in a Numerical Example}
\label{app:dp-benchmark}

We use a synthetic task stream to measure how closely \algname{} approaches the optimal expected solved count. The design preserves model choice and retries under a common budget, with discrete outcomes that make a dynamic program (DP) tractable. We vary cost uncertainty, difficulty persistence, and budget while holding the remaining primitives fixed. All parameters are set before testing.

\subsection{Experimental Design}
\label{app:dp-model}

Each stream contains 1,000 tasks and allows at most $H=4$ calls per task. The three models, indexed by $a\in\{L,M,S\}$, are listed in order of increasing success probability and mean output cost. A task's observed type $x\in\{E,D\}$ is easier ($E$) or more difficult ($D$). Types are drawn independently with equal probability. Hidden difficulty is good ($g$) with prior probability $\pi_x$ and bad ($b$) otherwise, where $\pi_E=0.75$ and $\pi_D=0.25$.

Let $p_{a,g}$ and $p_{a,b}$ denote model $a$'s success probabilities conditional on good and bad difficulty. Given the hidden state and selected model, each call uses a fresh Bernoulli draw. The first-call success probability for type $x$ is $p_a(x)=\pi_xp_{a,g}+(1-\pi_x)p_{a,b}$. Table~\ref{tab:dp-primitives} gives these probabilities and two output-length distributions. Lengths are deterministic in the fixed setting. In the tail setting, parentheses give the probabilities of the listed lengths.

\begin{table}[!htbp]
\centering
\caption{Success probabilities and output lengths in the synthetic environment.}
\label{tab:dp-primitives}
\small
\begin{tabular}{lrrrrll}
\toprule
Model & $p_{a,g}$ & $p_{a,b}$ & $p_a(E)$ & $p_a(D)$ & Fixed length & Tail length \\
\midrule
$L$ & 0.80 & 0.20 & 0.650 & 0.350 & 2 & 2 \\
$M$ & 0.90 & 0.40 & 0.775 & 0.525 & 3 & 2 (0.8), 7 (0.2) \\
$S$ & 0.95 & 0.55 & 0.850 & 0.650 & 4 & 2 (0.8), 12 (0.2) \\
\bottomrule
\end{tabular}
\end{table}

Hidden difficulty is either redrawn independently from the type-specific prior at every attempt, labeled iid in the tables, or held fixed within a task, labeled persistent. Output length is independent of difficulty and success. Crossing these two difficulty settings with the two length distributions and initial budgets $B\in\{2{,}000,5{,}000\}$ gives eight environments.

Before attempt $h$, let $u$ be the total output length of the current task's $h-1$ failed calls. A new call with output length $\ell$ costs $c=h+u+\ell$. Input and output have unit prices, and lengths and costs use synthetic units. Both policies make a first call whenever the balance is positive. After failure, they may call any model or abandon the task. Success, abandonment, or the fourth call ends the task. A call that reaches or crosses the budget is completed and credited before service stops, as in Section~\ref{sec:stopping}.

DP observes the same task type, call history, and budget ledger as \algname{}, but also knows the outcome distributions. Neither policy observes hidden difficulty or future responses. We retain the \algname{} learners, allocation rule, dual and pacing updates, and retry gate, with cost-label resolution $K=2$, tail tolerance $\delta_c=0.1$, and fixed controller defaults. Synthetic features replace embeddings. Let $f_a$ count failures of model $a$ on the current task, so $h=1+\sum_{a\in\{L,M,S\}}f_a$. The unprojected feature is
\[
[1,\1\{x=E\},\1\{x=D\},f_L/4,f_M/4,f_S/4,u/36,(1+u+\sum_a f_a)/40,\1\{h>1\}],
\]
where $\1\{\cdot\}$ is the indicator function and 4, 36, and 40 are fixed normalization constants. The shared reward component projects these nine coordinates to four dimensions.

Each environment uses 20 test seeds shared across policies and three equally weighted warm initializations, each based on 32 tasks outside the operating budget. Warm logs are reused across budgets, giving 160 DP and 480 \algname{} runs overall. Only selected responses are revealed. Confidence intervals (CIs) use 10,000 bootstrap resamples of the 20 test-seed blocks, retaining all three initializations within each block.

Observed type provides an imperfect difficulty signal, and persistent difficulty makes failures informative for later attempts. The input cost $h+u$ includes the initial prompt, retained failed outputs, and overhead from retries. The length settings vary uncertainty at a fixed mean output cost per call for each model, while the budgets vary resource scarcity. These settings reflect the general characteristics of LLM calls and routing.

\subsection{Dynamic Program}
Before revealing the next task's type, let $V(n,b)$ be the optimal expected additional solved count with $n$ tasks and balance $b$ remaining. Set $V(n,b)=0$ if $n=0$ or $b\le0$. Once the current type $x$ is observed, the state is $z=(n,b,x,\boldsymbol f,u)$, where $n$ includes the current task, $\boldsymbol f=(f_L,f_M,f_S)$ records its model-specific failure counts, and $u$ is its accumulated failed-output length. Let $W(z)$ be the optimal expected additional solved count conditional on this state. The next attempt is $h=1+\sum_a f_a$. Under persistent difficulty, let $\pi(z)$ denote the posterior probability of good difficulty, given the observed failures:
\begin{equation}
\pi(z)=\frac{\pi_x\prod_a(1-p_{a,g})^{f_a}}
{\pi_x\prod_a(1-p_{a,g})^{f_a}+(1-\pi_x)\prod_a(1-p_{a,b})^{f_a}}.
\label{eq:dp-posterior}
\end{equation}
For independent difficulty, $\pi(z)=\pi_x$. Write \mbox{$p_a(z)=\pi(z)p_{a,g}+[1-\pi(z)]p_{a,b}$} for model $a$'s next-call success probability at state $z$. Let $\nu_a(\ell)$ be the probability that model $a$ returns output length $\ell$, and let $\boldsymbol e_a$ be the unit vector that increments its failure count. Denote by $Q^*(z,a)$ the expected additional solved count from calling $a$ and acting optimally thereafter, and by $F(z,a,\ell)$ the optimal continuation value if that call fails. Summing over the possible output lengths gives
\begin{align}
Q^*(z,a)&=\sum_\ell\nu_a(\ell)\left[p_a(z)\{1+V(n-1,b-c)\}+[1-p_a(z)]F(z,a,\ell)\right],
\label{eq:dp-call}\\
F(z,a,\ell)&=\begin{cases}
W(n,b-c,x,\boldsymbol f+\boldsymbol e_a,u+\ell),&h<H,\ b-c>0,\\
V(n-1,b-c),&\text{otherwise},
\end{cases}
\label{eq:dp-failure}
\end{align}
Here $c=h+u+\ell$ is the realized call cost. Abandonment is available when $h>1$ and has action value $Q^*(z,\mathrm{abandon})=V(n-1,b)$. The value $W(z)$ is the maximum action value over the available calls and, when permitted, abandonment. A new task starts with failure-count vector $\boldsymbol0$ and accumulated output length zero. Averaging over its type gives
\begin{equation}
V(n,b)=\tfrac12\sum_{x\in\{E,D\}}W(n,b,x,\boldsymbol0,0).
\label{eq:dp-arrival}
\end{equation}
The recursion credits successful budget-crossing calls with zero continuation and permits final-task retries.

We define the Bellman loss as $D(z,a)=W(z)-Q^*(z,a)\ge0$. It measures the chosen action's loss against the best action at that state, with optimal continuation. Here $a$ can be either a model call or abandonment. Let $\pi$ be a nonanticipating routing policy, whose decisions use only information revealed before each decision, and let $\mathcal D_\pi$ index all decisions it visits, including abandonment. Write $z_j$ and $a_j$ for the state and action at decision $j$, and $\textsc{Solved}$ for the total number of tasks solved in the stream. For a fixed warm log, summing Bellman equations gives
\begin{equation}
V(1{,}000,B)-\E_\pi[\textsc{Solved}]
=\E_\pi\!\left[\sum_{j\in\mathcal D_\pi}D(z_j,a_j)\right],
\label{eq:dp-performance-loss}
\end{equation}

The expectation $\E_\pi$ averages task types and outcomes under policy $\pi$. Sampled losses separate first calls and retries.

\subsection{Results and Interpretation}

Let $V^*=V(1{,}000,B)$ denote DP's optimal expected solved count for the full stream. Table~\ref{tab:scaled-dp-results} reports solved counts and action agreement. Agreement means a value within $10^{-8}$ of the best action at the same \algname{}-visited state, including abandonment. Run-level fractions receive equal weight.

\begin{table}[!htbp]
\centering
\caption{Solved tasks and agreement with DP.}
\label{tab:scaled-dp-results}
\footnotesize
\setlength{\tabcolsep}{4pt}
\begin{tabular}{llrrrr}
\toprule
$B$ & Cost / difficulty & $V^*$ & ADP-HT [95\% CI] & ADP/$V^*$ & Agreement \\
\midrule
2,000 & Fixed / iid & 336.30 & 309.80 [306.47, 313.73] & 92.12\% & 31.32\% \\
2,000 & Fixed / persistent & 336.28 & 315.28 [312.97, 317.45] & 93.76\% & 33.78\% \\
2,000 & Tail / iid & 336.32 & 312.52 [308.10, 317.42] & 92.92\% & 34.07\% \\
2,000 & Tail / persistent & 336.31 & 315.48 [311.05, 319.92] & 93.81\% & 33.70\% \\
\midrule
5,000 & Fixed / iid & 763.27 & 741.98 [734.52, 749.47] & 97.21\% & 51.93\% \\
5,000 & Fixed / persistent & 750.76 & 736.97 [731.97, 741.93] & 98.16\% & 64.14\% \\
5,000 & Tail / iid & 767.50 & 735.55 [728.22, 742.65] & 95.84\% & 51.12\% \\
5,000 & Tail / persistent & 754.77 & 729.80 [723.35, 736.05] & 96.69\% & 59.66\% \\
\bottomrule
\end{tabular}
\par\smallskip
\begin{minipage}{.98\linewidth}\footnotesize
ADP-HT gives simulated means and 95\% test-seed-block intervals conditional on three fixed warm logs.
\end{minipage}
\end{table}

Ratios for individual warm logs range from $89.7\%$ to $100.4\%$, comparing simulated means with an expectation. All eight paired DP-minus-\algname{} mean gaps have positive 95\% intervals.

\label{app:dp-llm-interpretation}
Cumulative Bellman loss links decision differences to the expected reward gap in \eqref{eq:dp-performance-loss}. We examine budget use to explain where this loss arises.

\begin{table}[!htbp]
\centering
\caption{Behavior in the fixed-length independent environment.}
\label{tab:dp-behavior-example}
\small
\begin{tabular}{lrrrr}
\toprule
& \multicolumn{2}{c}{$B=2{,}000$} & \multicolumn{2}{c}{$B=5{,}000$} \\
\cmidrule(lr){2-3}\cmidrule(lr){4-5}
Measure & DP & ADP-HT & DP & ADP-HT \\
\midrule
Tasks served & 573.15 & 436.03 & 999.85 & 999.83 \\
First-call success among served tasks (\%) & 59.23 & 71.51 & 67.53 & 71.41 \\
First-call successes & 339.50 & 309.80 & 675.20 & 714.02 \\
Retry calls & 0.00 & 0.00 & 122.40 & 41.87 \\
Retry successes & 0.00 & 0.00 & 95.60 & 27.97 \\
Solved tasks & 339.50 & 309.80 & 770.80 & 741.98 \\
Calls / solution & 1.69 & 1.41 & 1.46 & 1.41 \\
Cost / solution & 5.91 & 6.47 & 6.50 & 6.73 \\
\bottomrule
\end{tabular}
\par\smallskip
\begin{minipage}{.98\linewidth}\footnotesize
Means use each policy's own trajectories. Rates and ratios are calculated per run before averaging. Costs and calls include failures. DP counts are simulated. Table~\ref{tab:scaled-dp-results} uses the optimal expectation.
\end{minipage}
\end{table}

\begin{figure}[H]
\centering
\setlength{\abovecaptionskip}{5pt}
\includegraphics[width=.92\linewidth]{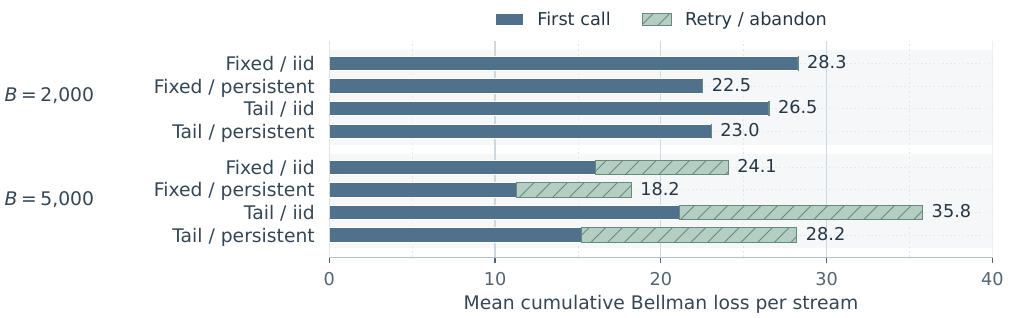}
\caption{Mean Bellman loss per stream under \algname{} by decision type.}
\label{fig:dp-value-behavior}
\end{figure}

\algname{} attains $92.12\%$ to $98.16\%$ of DP's optimal expected reward despite action agreement of only $31.32\%$ to $64.14\%$. Their different allocation patterns help interpret this result. First, under tight budgets, \algname{} favors the stronger model $S$, while DP mainly uses $L$ and $M$. In the fixed-length independent setting, its higher first-call success rate comes with greater cost per solution and fewer tasks served (Table~\ref{tab:dp-behavior-example}). Second, at the larger budget with fixed lengths, \algname{} retries less often than DP under independent difficulty but more often under persistent difficulty, indicating that its recovery decisions vary with the failure process. Finally, first calls account for nearly all sampled Bellman loss under tight budgets and $54\%$ to $67\%$ at the larger budget, leaving a larger share of the loss to retry and abandonment decisions (Figure~\ref{fig:dp-value-behavior}). These patterns also have practical relevance. Cascading accumulates the costs of successive calls \citep{dekoninck2025cascade}, and studies of conversations with progressively revealed instructions find that LLMs can retain earlier incorrect assumptions \citep{iclr2026-llm-lost}. Where failed outputs remain in context and hinder correction, a stronger first call may avoid repeated errors and growing input costs. Several cheap retries may then cost more than a single expensive call. Thus, \algname{}'s tendency to invest in initial success may be useful in real LLM routing scenarios.

\section{Tail Costs and the Value of Future Service}
\label{app:tail-routing}

We specialize Section~\ref{sec:model} to identical tasks, known distributions, and one call per task. We compare the future service protected by a predictable call with the extra budget it consumes. Throughout this section, \emph{backlog} means the number of unfinished tasks, including the current task; the initial backlog is $T$.

\subsection{Model and Optimal Call Sequence}
\label{app:tail-model}

The predictable model $P$ has cost $c$ and success probability $u$. The risky model $R$ has success probability $v$ and cost
\begin{equation}
\begin{aligned}
C_R&=\begin{cases}
\ell,&\text{with probability }q,\\
d,&\text{with probability }\varepsilon,
\end{cases}
\qquad q=1-\varepsilon,\\
&0<\ell<c<d,\quad d\ge B,\quad 0<u<v\le1,\quad 0<\varepsilon<1.
\end{aligned}
\label{eq:tail-primitives}
\end{equation}
Each call's success and cost are independent of the past, although they may be correlated within the call. A high-cost response always ends service and receives credit if successful under \eqref{eq:tau}.

Let $V_t(b)$ be the optimal expected number of solved tasks with $t$ tasks remaining and remaining budget $b$. Set $V_0(b)=0$ and $V_t(b)=0$ for $b\le0$. For $0<b\le B$,
\begin{equation}
V_t(b)=\max\{u+V_{t-1}(b-c),\ v+qV_{t-1}(b-\ell)\}.
\label{eq:tail-bellman}
\end{equation}
The immediate risky reward is $v$, including the high-cost response. For $b>c$, let $Q_P^*(t,b)$ and $Q_R^*(t,b)$ be the optimal expected numbers of solved tasks when the first call uses $P$ and $R$, respectively, with optimal choices thereafter. Their difference is
\begin{equation}
Q_P^*(t,b)-Q_R^*(t,b)
=-(v-u)+\varepsilon V_{t-1}(b-c)
-q\{V_{t-1}(b-\ell)-V_{t-1}(b-c)\}.
\label{eq:tail-value-decomposition}
\end{equation}
The last bracket is nonnegative because additional budget cannot reduce the optimum. A predictable call reduces the immediate success probability by $v-u$ and uses $c-\ell$ more budget than a low-cost risky call. Its benefit is the future service protected from termination.

Write $G(k)=\sum_{i=0}^{k-1}q^i=(1-q^k)/\varepsilon$, with $G(0)=0$. Thus $vG(k)$ is the expected number of solved tasks from at most $k$ risky calls, including the high-cost response. We call an initial sequence of predictable calls the \emph{predictable prefix} and the following risky calls the \emph{risky suffix}.
\begin{proposition}[Predictable calls precede risky calls]
\label{prop:tail-prefix}
Under \eqref{eq:tail-primitives}, consider a predictable prefix of length $m$ and define
\[
m_{\max}=\min\{T-1,\lceil B/c\rceil-1\},\qquad
k_m=\min\{T-m,\lceil(B-mc)/\ell\rceil\}.
\]
Here $k_m$ is the maximum number of risky calls that can follow this prefix. Then
\begin{equation}
V_T(B)=\max_{m=0,\ldots,m_{\max}}\{mu+vG(k_m)\}.
\label{eq:tail-prefix}
\end{equation}
An optimal policy makes a maximizing number $m$ of predictable calls first and then uses $R$ on all remaining tasks until service stops to banlance the rewards and the risk.
\end{proposition}

\begin{proof}
Backward induction gives a deterministic optimal policy whose choices depend only on the current state. Every risky call that does not end service costs $\ell$, so the model choices before any high-cost response form a deterministic sequence. Past successes give no information about future tasks. A predictable call on the final task or one that exhausts the budget can be replaced by $R$: the immediate expected reward increases, and the expected reward from later tasks remains nonnegative.

Any adjacent pair $R,P$ in an optimal sequence therefore starts with remaining budget greater than $\ell+c$. Swapping it to $P,R$ leaves the same state after a low-cost risky call. The expected rewards through the pair change from $v+qu$ to $u+v$, a strict gain of $\varepsilon u$. Thus all predictable calls precede all risky calls. This predictable prefix must leave positive budget and at least one task. Its risky suffix admits at most $k_m$ calls, with call $i$ reached with probability $q^{i-1}$. This proves \eqref{eq:tail-prefix} over all nonanticipating policies.
\end{proof}

\subsection{Backlog and the Optimal First Model}
\label{app:tail-backlog}

\begin{theorem}[A reversal as backlog increases]
\label{thm:tail-backlog}
Suppose $c<B\le2c$. Put $z=\lceil B/\ell\rceil$, $j=\lceil(B-c)/\ell\rceil$, and $a=u/v$. An optimal policy either always chooses $R$ or chooses $P$ once and then uses $R$. For $T\ge2$, their values are
\begin{align}
F_0(T)&=vG(\min\{T,z\}),&
F_1(T)&=u+vG(\min\{T-1,j\}),\label{eq:tail-two-values}\\
D_T&=G(\min\{T,z\})-G(\min\{T-1,j\}).&&\label{eq:tail-threshold}
\end{align}
Then $V_T(B)=\max\{F_0(T),F_1(T)\}$. Model $P$ is optimal first exactly when $a\ge D_T$, where
\begin{equation}
D_T=\begin{cases}
q^{T-1},&1\le T\le j+1,\\
q^jG(T-j),&j+1\le T\le z,\\
q^jG(z-j),&T\ge z.
\end{cases}
\label{eq:tail-backlog-threshold}
\end{equation}
Strict comparisons give a unique first model and equality gives a tie. At $T=1$, $R$ is uniquely optimal. A strict preference for $R$, then $P$, then $R$ as $T$ increases occurs exactly when
\begin{equation}
q^j<a<q^jG(z-j).
\label{eq:tail-reversal-condition}
\end{equation}
\end{theorem}

\begin{proof}
The budget restriction permits at most one predictable call that does not exhaust the budget, so Proposition~\ref{prop:tail-prefix} gives $F_0$ and $F_1$. Their difference is $v(a-D_T)$. The adjacent-swap argument also implies that a globally optimal policy starting with $R$ uses no later predictable call. Hence this comparison identifies the optimal first model.

For $T\le j+1$, subtracting $G(T-1)$ from $G(T)$ gives $q^{T-1}$. The number of risky calls following $P$ stops increasing at $j$, giving the second expression. Both expected values stop increasing once $T\ge z$. The expressions agree at their endpoints. Thus $D_T$ decreases to $q^j$, then increases to $q^jG(z-j)$ if $z-j>1$, and stays constant thereafter. Since $a<1$, a strict middle preference for $P$ and a strict eventual preference for $R$ are equivalent to \eqref{eq:tail-reversal-condition}.
\end{proof}

A short backlog leaves little future service to protect. An intermediate backlog makes avoiding termination valuable. With a long backlog, the low-cost risky calls displaced by predictable service can restore the preference for $R$.

\begin{figure}[!htbp]
\centering
\includegraphics[width=.9\linewidth]{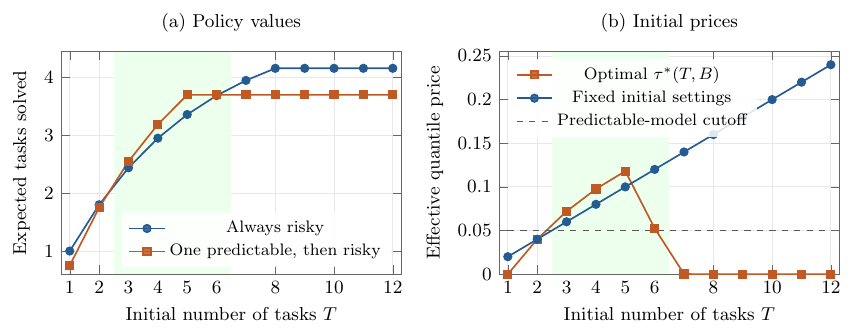}
\caption{Exact policy values in panel (a) and effective cost prices in panel (b). Shading marks backlogs where $P$ is optimal first. Each backlog is a separate initial state with the same budget and model distributions.}
\label{fig:tail-backlog}
\end{figure}

For an exact example, take $(\ell,c,d,\varepsilon,u,v,B)=(1,4,9,0.2,0.75,1,7.5)$. Model $R$ has both a higher success probability and a lower mean cost, $2.6$ compared with $4$. Here $z=8$, $j=4$, and $0.4096<0.75<1.2091392$. The optimal first model is $R$ for $T=1,2$, $P$ for $T=3,4,5,6$, and $R$ for $T\ge7$. At $T=5$, a predictable first call followed by risky calls yields $3.702$ expected successes, compared with $3.3616$ under always choosing $R$, a $10.1\%$ increase. Figure~\ref{fig:tail-backlog} shows the exact values.

\subsection{General Budgets}
\label{app:tail-general-budgets}
When $c<B\le2c$, at most one predictable call can leave positive budget for another call. For general $B>c$, several predictable calls may be optimal. Proposition~\ref{prop:tail-prefix} still characterizes the predictable-first backlog region and when two reversals occur. Even with multiple predictable calls, that region remains an interval rather than alternating arbitrarily between $P$ and $R$.

\begin{theorem}[Backlog intervals for general budgets]
\label{thm:tail-general-backlog}
Suppose $B>c$, and retain $a=u/v$. Define
\[
m_B=\left\lceil B/c\right\rceil-1,\qquad
z_m=\left\lceil(B-mc)/\ell\right\rceil\quad(0\le m\le m_B),\qquad
T^*=1+z_1.
\]
Thus $z_m$ is the number of risky calls that fit after $m$ predictable calls, including the crossing response. For $T\ge2$, model $P$ is an optimal first choice exactly when
\begin{equation}
\begin{cases}
a\ge q^{T-1},&T\le T^*,\\[2pt]
\displaystyle\max_{m=1,\ldots,m_B}\{ma+G(z_m)\}
\ge G(\min\{T,z_0\}),&T\ge T^*.
\end{cases}
\label{eq:tail-general-first}
\end{equation}
The two cases give the same comparison at $T=T^*$. Strict inequalities give a unique first model, and equality gives a tie. The predictable-first backlogs form an interval. A strict preference for $R$, then $P$, then $R$ as $T$ increases occurs exactly when
\begin{equation}
q^{z_1}<a<
\min_{m=1,\ldots,m_B}\frac{G(z_0)-G(z_m)}{m}.
\label{eq:tail-general-reversal}
\end{equation}
\end{theorem}

\begin{proof}
For $0\le m\le m_B$, the value of $m$ predictable calls followed by risky calls is
\[
F_m(T)=mu+vG(\min\{T-m,z_m\})
=v\left[ma+G(\min\{T-m,z_m\})\right].
\]
The expected value of the risky suffix stops increasing at $T=m+z_m$. Since
\[
m+z_m=\left\lceil B/\ell-m(c/\ell-1)\right\rceil,
\]
these thresholds are nonincreasing in $m$, and $T^*\le z_0$.

If $T\le T^*$, then $F_0(T)=vG(T)$ and $F_1(T)=u+vG(T-1)$. Every feasible positive prefix satisfies
\[
\frac{F_m(T)-F_0(T)}{v}
\le ma+G(T-m)-G(T)
=ma-\sum_{i=T-m}^{T-1}q^i
\le m(a-q^{T-1}).
\]
Thus no positive prefix strictly improves on always choosing $R$ when $a<q^{T-1}$. The one-call prefix strictly improves when $a>q^{T-1}$ and ties when $a=q^{T-1}$. This gives the first case in \eqref{eq:tail-general-first}.

If $T\ge T^*$, every positive prefix is feasible and the expected value of its risky suffix has stopped increasing. The best value divided by $v$ among these prefixes is therefore $\max_m\{ma+G(z_m)\}$, which gives the second case. The adjacent-swap argument in Proposition~\ref{prop:tail-prefix} converts these policy comparisons into first-model comparisons.

Before $T^*$, the threshold $q^{T-1}$ strictly decreases. After $T^*$, the best value among prefixes with $m\ge1$ is fixed, while the always-risky value weakly increases. Thus the predictable-first backlogs form an interval. At $T=T^*$, the comparison is $a$ versus $q^{z_1}$, so $P$ is strictly optimal first for some $T$ exactly when $a>q^{z_1}$. A strict eventual return to $R$ occurs exactly when $\max_m\{ma+G(z_m)\}<G(z_0)$, giving \eqref{eq:tail-general-reversal}. Model $R$ is uniquely optimal at $T=1$ because $u<v$.
\end{proof}

When $c<B\le2c$, $m_B=1$, recovering Theorem~\ref{thm:tail-backlog}.

\subsection{Quantile Scores and Expected Completions}
\label{app:tail-quantile}

Let $\mathcal Q_a(1-\delta_c)=\inf\{x\mid\Pr(C_a\le x)\ge1-\delta_c\}$. For $\delta_c<\varepsilon$, the two quantiles are $\mathcal Q_P=c$ and $\mathcal Q_R=d$. For $\delta_c\ge\varepsilon$, the risky quantile is $\ell$, including equality under this convention. Write $p_P=u$ and $p_R=v$. Throughout this subsection, each score keeps its success probabilities, cost estimates, and effective price $\tau$ fixed as $T$ varies.

\begin{proposition}[Help and harm from a fixed quantile penalty]
\label{prop:tail-fixed-quantile}
Suppose $q\ell+\varepsilon d\le c$, $\delta_c<\varepsilon$, and $\tau>(v-u)/(d-c)$. The fixed score $p_a-\tau\E C_a$ always selects $R$, whereas $p_a-\tau\mathcal Q_a(1-\delta_c)$ always selects $P$. Writing $k=\lceil B/c\rceil$ and $z=\lceil B/\ell\rceil$, their values are
\begin{equation}
J_{\rm mean}(T)=vG(\min\{T,z\}),\qquad
J_{\rm quant}(T)=u\min\{T,k\}.
\label{eq:tail-fixed-values}
\end{equation}
The quantile score improves expected successes if and only if $(u/v)\min\{T,k\}>G(\min\{T,z\})$. Strict help at intermediate backlogs and harm at $T=1$ and all $T\ge z$ occur exactly when
\begin{equation}
G(k)<(u/v)k<G(z).
\label{eq:tail-fixed-condition}
\end{equation}
\end{proposition}

\begin{proof}
The assumptions give strict score preferences. The call counts in \eqref{eq:tail-fixed-values} include the crossing response. Up to $T=k$, $G(T)/T$ strictly decreases because it averages $1,q,\ldots,q^{T-1}$. An improvement from the quantile score can therefore begin only once. From $k$ to $z$, the predictable value is constant and the risky value increases. Both are constant afterwards. Any strict improvement must include $T=k$. Combining its inequality with strict harm after $z$ gives \eqref{eq:tail-fixed-condition}.
\end{proof}

Thus the model choices under both fixed scores are constant, but their ranking in expected solved tasks can change with the backlog. The changing value of future service drives this difference.

Even the mean and a single upper quantile together need not identify the optimal first model. Fix $c=3$, $u=0.85$, $v=1$, $B=7$, $T=3$, and $\delta_c=0.025$. In instance A, $R$ costs $0.5$ with probability $0.8$ and $10$ otherwise. In instance B, it costs $2$ with probability $0.95$ and $10$ otherwise. Both have mean $2.4$ and $0.975$-quantile $10$.

Without a high risky cost, any two calls cost at most $6<B$, so all three tasks are attempted unless a high risky cost occurs. Let $W_t$ be the optimal expected number of solved tasks with $t$ tasks remaining. The recursion becomes $W_t=\max\{u+W_{t-1},v+qW_{t-1}\}$, with $W_0=0$. Instance A has optimal sequence $P,P,R$ and value $2.70$. Instance B optimally uses $R$ throughout and has value $2.8525$. Thus the same mean and quantile summaries give opposite optimal first choices. At $\tau=0.03$, fixed quantile pricing selects $P$ throughout and earns $2.55$ in both instances. It improves on always choosing $R$ by $0.11$ in A and loses $0.3025$ in B. The difference is the probability of losing later service. For a fixed model-choice sequence, increasing $d$ further above $B$ changes neither when service stops nor the number of solved tasks. It still changes both fixed scores.

\subsection{Implications for the Effective Cost Price}
\label{app:tail-controller}

In \eqref{eq:adpht-score}, the effective price is $\tau_s=\lambda_s/b_s^{\rm task}$. Consider initial decisions with the known success probabilities and cost quantiles. Hold these predictions, $\lambda_0\ge0$, $\phi_0>0$, and $B$ fixed while varying only $T$. Equation~\eqref{eq:local-budget-task} gives
\begin{equation}
\tau_0(T)=\frac{\lambda_0}{\max\{\phi_0B/T,\epsilon\}},\qquad
Q_0(P)-Q_0(R)=u-v+\tau_0(T)(d-c)
\label{eq:tail-initial-price}
\end{equation}
when $\delta_c<\varepsilon$. This difference is nondecreasing in $T$, so fixed initialization cannot reproduce both reversals in Theorem~\ref{thm:tail-backlog}. For $\delta_c\ge\varepsilon$, $R$ wins at every nonnegative price.

When $c<B\le2c$, there is also an explicit price that implements the optimal choices. At a state with $t\ge2$ and $c<b\le2c$, define
\begin{align}
D_t(b)&=G\!\left(\min\{t,\lceil b/\ell\rceil\}\right)
-G\!\left(\min\{t-1,\lceil(b-c)/\ell\rceil\}\right),\label{eq:tail-state-threshold}\\
\tau^*(t,b)&=\frac{v[1-D_t(b)]_+}{d-c}.
\label{eq:tail-optimal-price}
\end{align}
Set $\tau^*(t,b)=0$ for states with $b\le c$ or $t=1$. For any fixed $\delta_c<\varepsilon$, maximizing $p_a-\tau^*(t,b)\mathcal Q_a(1-\delta_c)$ gives an optimal policy under $B\le2c$.

If $D_t(b)<1$, the score difference is $u-vD_t(b)$, as in Theorem~\ref{thm:tail-backlog}. Otherwise the price is zero and $R$ is optimal since $u<v$. Model $R$ also dominates when $b\le c$ or $t=1$. Backward induction establishes optimality. For fixed $b$, the price first weakly increases and then weakly decreases with $t$. It gives the optimal comparison between the two models and is a coefficient in the score, not necessarily the value of adding one unit of budget.

Equation~\eqref{eq:tail-optimal-price} makes the adaptation explicit through the remaining budget and backlog in $D_t(b)$ and the risky cost probability in $G$. In \algname{}, $b_s^{\rm task}$ contains the remaining budget per unfinished task, and $\lambda_s$ and $\phi_s$ respond to realized spending. The exact $\tau^*(t,b)$ is a structural benchmark for how the price should vary with the value of future service. \algname{}'s empirical updates allow its effective price to respond to related state variables, which leading to better performance in routing.

Figure~\ref{fig:tail-backlog}(b) uses $\lambda_0=0.15$ and $\phi_0=1$, giving $\tau_0(T)=0.02T$ because $\epsilon$ does not affect the plotted values. This initialized price crosses the cutoff $(v-u)/(d-c)=0.05$ once, whereas $\tau^*(T,7.5)$ crosses it twice.

\end{document}